%% file: iclr2027_conference.tex
\documentclass{article}
\usepackage{iclr2027_conference,times}   

\usepackage{amsmath,amssymb,amsfonts}
\usepackage{graphicx}
\usepackage{booktabs}      
\usepackage{multirow}
\usepackage{array}
\usepackage{subcaption}     
\usepackage{xcolor}
\usepackage{xspace}
\usepackage{microtype}     
\usepackage{url}
\usepackage{tabularx}
\graphicspath{{figures/}}
\usepackage{placeins}
\usepackage{hyperref}                          
\usepackage[capitalise,noabbrev]{cleveref}     
\input{math_commands.tex}

\input{macros.tex}

\title{MacJEPA: Missingness-Robust Audio-Visual Recognition from Untrimmed Egocentric Videos}

\author{
Anonymous \\
Anonymous Institution \\
\texttt{anonymous@example.com}
}
\author{Souptik Sen$^{1}$, Zahra Ahmadi$^{1,2}$\\
$^1$Peter L. Reichertz Institute for Medical Informatics, Hannover Medical School, Germany\\
$^2$Lower Saxony Center for Artificial Intelligence and Causal Methods in Medicine (CAIMed), Hannover, Germany\\
\texttt{\{Sen.Souptik,Ahmadi.Zahra\}@mh-hannover.de}
}

\newif\ifworkinprogress
\workinprogresstrue 

\ifworkinprogress

\else

\fi

\iclrfinalcopy   

\begin{document}

\maketitle

\begin{abstract}
\input{sections/00_abstract}
\end{abstract}

\input{sections/01_introduction}

\input{sections/02_related_work}
\input{sections/03_method}
\input{sections/04_experiments}
\input{sections/05_conclusion}

\input{sections/07_statements}

\bibliographystyle{iclr2027_conference}
\bibliography{iclr2027_conference}

\clearpage
\appendix
\input{sections/06_appendix}

\end{document}

%% file: math_commands.tex
\usepackage{amsmath,amsfonts,bm}

\def\eqref#1{equation~\ref{#1}}

\def\1{\bm{1}}

\def\vtheta{{\bm{\theta}}}

\def\vc{{\bm{c}}}

\def\ve{{\bm{e}}}

\def\vu{{\bm{u}}}

\def\vx{{\bm{x}}}

\def\vz{{\bm{z}}}

\def\mM{{\bm{M}}}

\def\mX{{\bm{X}}}

\def\mZ{{\bm{Z}}}

\DeclareMathAlphabet{\mathsfit}{\encodingdefault}{\sfdefault}{m}{sl}
\SetMathAlphabet{\mathsfit}{bold}{\encodingdefault}{\sfdefault}{bx}{n}

%% file: macros.tex
\RequirePackage{amsmath}
\RequirePackage{amssymb}
\RequirePackage{xcolor}
\RequirePackage{xspace}

\newcommand{\EK}{Epic-Kitchens\xspace}

\newcommand{\ES}{Epic-Sounds\xspace}

\newcommand{\TIM}{TIM\xspace}

\providecommand{\vx}{\mathbf{x}}
\providecommand{\vu}{\mathbf{u}}
\providecommand{\vz}{\mathbf{z}}
\providecommand{\ve}{\mathbf{e}}
\providecommand{\vc}{\mathbf{c}}

\providecommand{\mX}{\mathbf{X}}
\providecommand{\mM}{\mathbf{M}}

\providecommand{\mZ}{\mathbf{Z}}

\providecommand{\vtheta}{\boldsymbol{\theta}}
\providecommand{\vpsi}{\boldsymbol{\psi}}

\newcommand{\Mset}{\mathcal{M}}
\newcommand{\vmod}{\mathrm{v}}
\newcommand{\amod}{\mathrm{a}}

\newcommand{\Nrow}{T}
\newcommand{\wsz}{w}
\newcommand{\Nwin}{K}
\newcommand{\Ncon}{n}

\newcommand{\fstride}{\Delta}
\newcommand{\fextent}{\omega}
\newcommand{\obsspan}{S}
\newcommand{\windursym}{\tau}

\newcommand{\featdim}[1]{D_{#1}}
\newcommand{\feat}[2]{\vx^{#1}_{#2}}
\newcommand{\featseq}[1]{\mX^{#1}}

\newcommand{\rowiv}[1]{\mathbf{t}_{#1}}

\newcommand{\qiv}[1]{\mathbf{t}_{#1}}

\newcommand{\Qset}{\mathcal{Q}}

\newcommand{\Qvalid}{\mathcal{Q}_{\mathrm{valid}}}
\newcommand{\Qhead}[1]{\mathcal{Q}_{#1}}

\newcommand{\Hset}{\mathcal{H}}

\newcommand{\qlabel}[1]{y_{#1}}

\newcommand{\ovl}{\delta}

\newcommand{\stem}[1]{g^{#1}}
\newcommand{\tmlp}{\phi}

\newcommand{\dcon}{d_{c}}
\newcommand{\dtim}{d_{t}}
\newcommand{\dmod}{d}

\newcommand{\memb}[1]{\ve^{#1}}
\newcommand{\mskemb}[1]{\ve^{#1}_{\varnothing}}
\newcommand{\clsv}[1]{\vc_{#1}}

\newcommand{\Mask}{\mM}

\newcommand{\Mel}[2]{M_{#1,#2}}

\newcommand{\pdrop}{\pi}

\newcommand{\assemble}{\mathcal{U}}

\newcommand{\ctok}[3]{\vu^{#1}_{#2,#3}}
\newcommand{\qtok}[1]{\vu_{#1}}

\newcommand{\enc}{f_{\vtheta}}
\newcommand{\tenc}{f_{\bar{\vtheta}}}
\newcommand{\pred}{h_{\vpsi}}
\newcommand{\emam}{\alpha}

\newcommand{\Zctx}{\mZ^{c}}
\newcommand{\Zqry}{\mZ^{q}}
\newcommand{\tZctx}{\bar{\mZ}^{c}}
\newcommand{\tZqry}{\bar{\mZ}^{q}}

\newcommand{\sg}[1]{\operatorname{sg}\!\left[#1\right]}
\newcommand{\lnz}[1]{\operatorname{LN}_{0}\!\left(#1\right)}

\newcommand{\Lce}{\mathcal{L}_{\mathrm{CE}}}
\newcommand{\Ljepa}{\mathcal{L}_{\mathrm{JEPA}}}
\newcommand{\Lvic}{\mathcal{L}_{\mathrm{VIC}}}
\newcommand{\Ltot}{\mathcal{L}}

\newcommand{\lctx}{\ell_{\mathrm{ctx}}}
\newcommand{\lqry}{\ell_{\mathrm{qry}}}

\newcommand{\uenc}[1]{b^{#1}}


%% file: sections/00_abstract.tex


Audio-visual models improve egocentric action recognition by exploiting
complementary cues, yet typically assume that both streams remain available at
inference. Existing missing-modality methods operate on trimmed, single-event
clips in which a stream is entirely present or absent, whereas real sensors fail and recover within long, untrimmed observations. We redefine egocentric modality
missingness as temporally localized sensor outages within untrimmed, multi-event
observations, with whole-clip absence as the limiting case. We introduce
\textbf{MacJEPA}, a missing-modality-robust
\textbf{Ma}sked-\textbf{c}ontext query
\textbf{JEPA} that recognizes visual actions and acoustic events from
supplied interval queries over audio-visual context. Window-local modality dropout simulates these sensor outages during training. MacJEPA further repurposes masking in JEPA from a self-supervised pretext into a supervised robustness objective, aligning masked and clean latent representations of both multimodal content tokens and the
task-conditioned queries. All objectives are optimized jointly with recognition in a single stage, requiring no test-time adaptation. Across Epic-Kitchens-100 and Epic-Sounds, a single checkpoint
remains competitive under complete input and consistently surpasses published
missing-modality baselines when either the dominant or auxiliary stream is
removed. MacJEPA thus unifies strong full-input recognition with temporal
missing-modality robustness in a single model operating on untrimmed multi-event
videos.

%% file: sections/01_introduction.tex

\section{Introduction}
\label{sec:intro}

Egocentric video captures activity from the actor's perspective, and understanding it underpins assistive systems, robotics, and augmented reality~\citep{damen2022rescaling,grauman2022ego4d}. Vision alone is a fragile signal in this setting; head motion, occlusion, and out-of-view interactions often hide evidence, and many egocentric actions remain acoustically identifiable when visual evidence is occluded or out of view.~\citep{kazakos2019epicfusion,huh2023epicsounds}. Audio-visual models exploit this complementarity and define the state of the art in egocentric action recognition through convolutional or transformer-based models ~\citep{kazakos2019epicfusion,kazakos2021mtcn,nagrani2021mbt,chalk2024tim}. Nevertheless, their gains are contingent on the availability of both streams at inference time, a premise that is often violated during deployment. For example, in real-world wearable devices, microphones or recordings can be suppressed for privacy, sensors can be duty-cycled to conserve power, or hardware can malfunction~\citep{gong2023mmgego4d,ramazanova2025mmt,ramazanova2025midl,santosvillafranca2026karmma}. The degradation is rarely graceful; losing one stream can push a multimodal model below a unimodal model trained on the surviving stream~\citep{ramazanova2025mmt}. This makes missing-modality resilience a deployment necessity rather than an optional refinement.

\begin{figure}[t]
\begin{center}
\includegraphics[width=\linewidth, height=7.7cm]{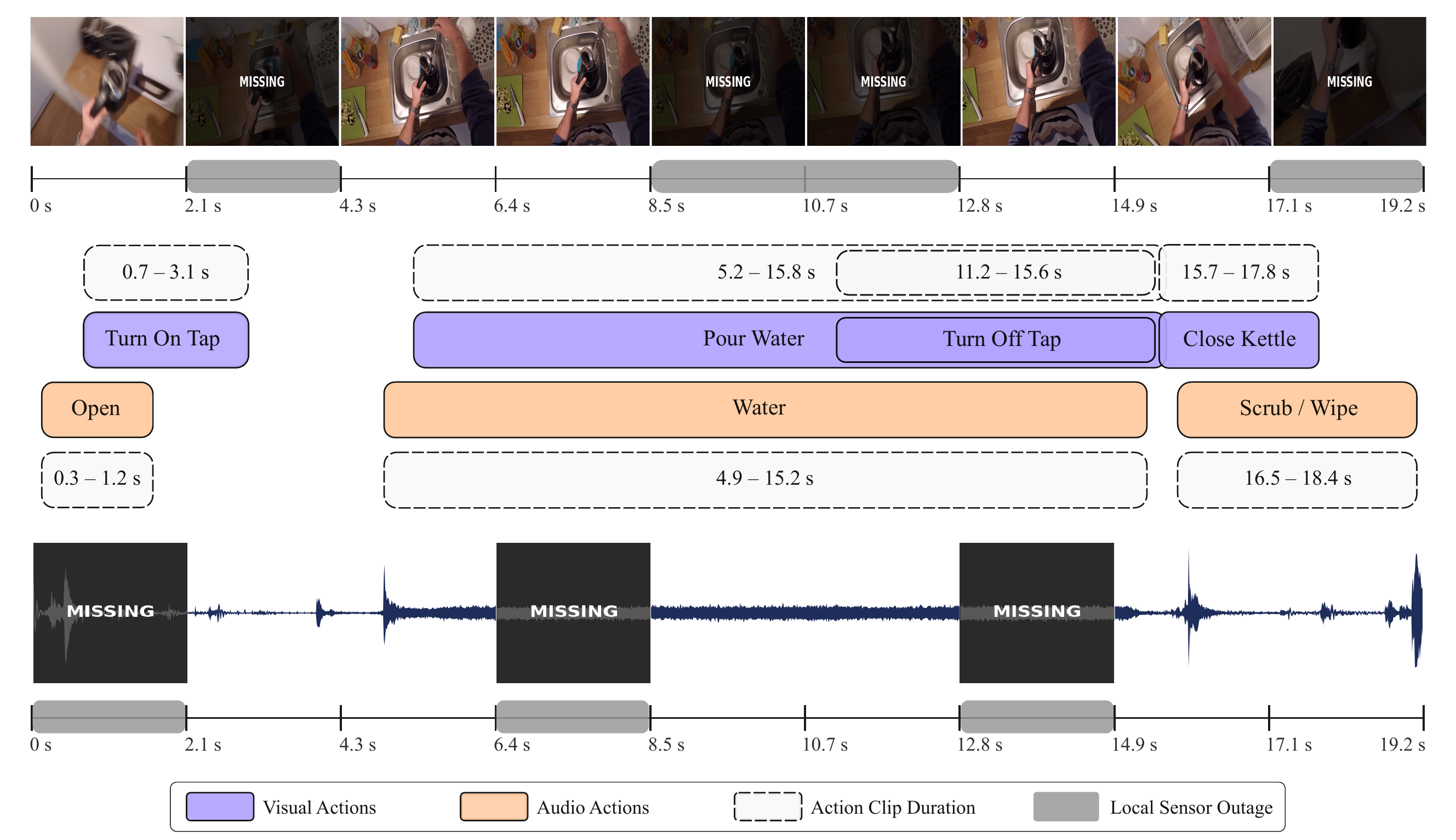}
\end{center}
\caption{\small Local sensor outages across overlapping multimodal events in a 19.2-second untrimmed observation.}
\label{fig:intro}
\end{figure}

Existing egocentric missing-modality methods remain tied to atomic clips centered on a
single annotated event, treating each stream as present or absent throughout the
sample~\citep{gong2023mmgego4d,ramazanova2025mmt,ramazanova2025midl,
santosvillafranca2026karmma,ma2022robust,lee2023prompting}. Missingness is consequently measured across clips, which implicitly align sensor outages with action boundaries. Even methods incorporating temporal context either aggregate neighboring action-centered clips or assume complete modalities over an untrimmed stream~\citep{kazakos2021mtcn,chalk2024tim}. Deployment grants no such action boundaries, requiring systems to process background, partial actions, and multiple temporally overlapping visual and acoustic events~\citep{damen2022rescaling,huh2023epicsounds,chalk2024tim}. Within such continuous observations, either sensor may disappear and return, interleaving clean and corrupted evidence at different intervals. Since overlapping actions share this context, one local outage can affect recognition of several events at different times and to varying degrees. \Cref{fig:intro} follows a plausible wearable-kitchen sequence where the wearer turns on the tap, pours water, turns it off, and closes the kettle, accompanied by the acoustic events \emph{Open}, \emph{Water}, and \emph{Scrub/Wipe}. Initial privacy muting removes acoustic evidence for \emph{Open} before audio returns, while a brief camera failure clips visual evidence for \emph{Turn On Tap} before video recovers. During the overlapping \emph{Pour Water}, \emph{Turn Off Tap}, and \emph{Water} events, audio is lost over 6.4--8.5\,s, video over 8.5--12.8\,s, and audio again over 12.8--14.9\,s, repeatedly changing which stream can support recognition. These local outages therefore cause multiple overlapping events to lose different degrees of evidence, revealing what atomic clip-level binary missingness cannot represent. The operative question is no longer which modality is absent from a clip, but which is absent \emph{when} and which events consequently lose distinguishable evidence. 

We introduce \textbf{MacJEPA}, a missing modality robust \textbf{Ma}sked-\textbf{c}ontext query \textbf{J}oint-\textbf{E}mbedding \textbf{P}redictive \textbf{A}rchitecture that recognizes
actions by querying supplied event intervals within untrimmed audio-visual
context. During training, MacJEPA independently drops each modality over short intervals within an untrimmed observation and substitutes them with learned mask vectors. Each query must preserve its prediction from surviving temporally aligned evidence, making the missingness in \Cref{fig:intro} a training condition rather than a deployment-time distribution shift. Canonical JEPA models train a masked online encoder to match clean Exponential Moving Average teacher targets in the latent space, providing an unexplored promising direction for this robustness machinery~\citep{baevski2022data2vec,assran2023ijepa,assran2025vjepa2}. Moreover, aligning masked multimodal context with clean targets in a shared latent space avoids reconstructing heterogeneous audio and visual signals~\citep{lei2025m3jepa}. Existing frameworks, however, use masking as a self-supervised pretext for transferable features rather than to model deployment failures
~\citep{assran2025vjepa2,teotia2026mjepa,robson2026avjepa,lei2025m3jepa}. MacJEPA repurposes JEPA's masked-to-clean consistency as a robustness objective optimized jointly with task supervision.

Our contributions are threefold. \textbf{(1)} We redefine egocentric modality missingness as an interval property of multi-event, untrimmed observations, with whole-clip binary absence as its limiting case. \textbf{(2)} We introduce interval-wise modality dropout with typed, time-stamped mask tokens, modeling local sensor outages and recoveries within one observation. \textbf{(3)} We reimagine JEPA as a task-supervised robustness objective that aligns masked and clean latent states
of multimodal content and task-conditioned queries. With complete input, \textbf{MacJEPA} achieves competitive accuracies of 54.0\% on \EK action and 58.1\% on \ES. Under total dominant-modality loss, it reaches 49.5\% \EK verb and 50.0\% \ES sound accuracy, surpassing the strongest published baseline by 4.2 and 9.3 points at this protocol-matched endpoint~\citep{ramazanova2025mmt}.

%% file: sections/02_related_work.tex
\section{Related Work}
\label{sec:related}

\textbf{Multimodal Egocentric Recognition under Missing Modalities.}
Audio complements egocentric vision because contact and state changes remain
audible when hands or objects leave the frame
~\citep{kazakos2019epicfusion,huh2023epicsounds}. Convolutional designs include
\textsc{TBN}, which fuses RGB, flow, and audio; \textsc{AVSlowFast}, which
couples fast audio with SlowFast vision; and \textsc{ASF}, which adapts this
design to acoustic recognition
~\citep{kazakos2019epicfusion,xiao2020avslowfast,kazakos2021asf}.
Transformers advance this domain by learning fusion through \textsc{MBT}'s bottleneck tokens, \textsc{MTCN}'s neighboring clips, and \textsc{M\&M Mix}'s multiple views,
while \textsc{TIM} queries both streams over one time-encoded untrimmed timeline
~\citep{nagrani2021mbt,kazakos2021mtcn,xiong2022mmmix,chalk2024tim}. These
strong baselines assume continuous access to both modalities, while
missing-modality methods remain sample-atomic. \textsc{MMG-Ego4D} evaluates
withheld modality subsets, \textsc{MMT} learns an absent-stream token, and
\textsc{KARMMA} distills a multimodal teacher into a dropout-trained student
supporting any modality combination
~\citep{gong2023mmgego4d,ramazanova2025mmt,santosvillafranca2026karmma}.
\textsc{MiDl} instead adapts at inference while distilling full-modality
predictions, whereas related work distills multimodal teachers into RGB-only
students~\citep{ramazanova2025midl,radevski2023multimodal}. Fusion search and
missing-aware prompting retain the same sample-wise binary premise
~\citep{ma2022robust,lee2023prompting}, measuring affected action-centered
samples rather than outage timing. MacJEPA unifies these lines by modeling
disappearance and recovery within shared, multi-event untrimmed context, with
clip-level absence as its limiting case.

\textbf{Multimodal JEPAs.}
JEPAs predict abstract targets rather than reconstruct raw inputs, avoiding
modality-specific low-level detail in heterogeneous settings
~\citep{assran2023ijepa,baevski2022data2vec,fei2023ajepa,lei2025m3jepa}.
\textsc{M3-JEPA} aligns modalities with a multi-gate mixture-of-experts
predictor, while \textsc{MJEPA} uses a shared encoder for within- and
cross-modal audio-video prediction
~\citep{lei2025m3jepa,teotia2026mjepa}. \textsc{AV-JEPA} aligns global
audio-visual and local unimodal views through early fusion and modality dropout,
whereas \textsc{VL-JEPA} predicts continuous text embeddings from vision
~\citep{robson2026avjepa,chen2026vljepa}. Across these methods, masking remains
a self-supervised pretext for transferable features rather than a deployment
failure, and robustness to incomplete inputs is not evaluated. \textsc{MJEPA}
leaves this direction to future work~\citep{teotia2026mjepa}. MacJEPA
repurposes masking to simulate sensor outages, optimizes it jointly with
supervision, and extends prediction from content tokens to task-conditioned
queries.

%% file: sections/03_method.tex
\section{Proposed Framework}
\label{sec:method}

\begin{figure}[t]
\begin{center}
\includegraphics[width=\linewidth]{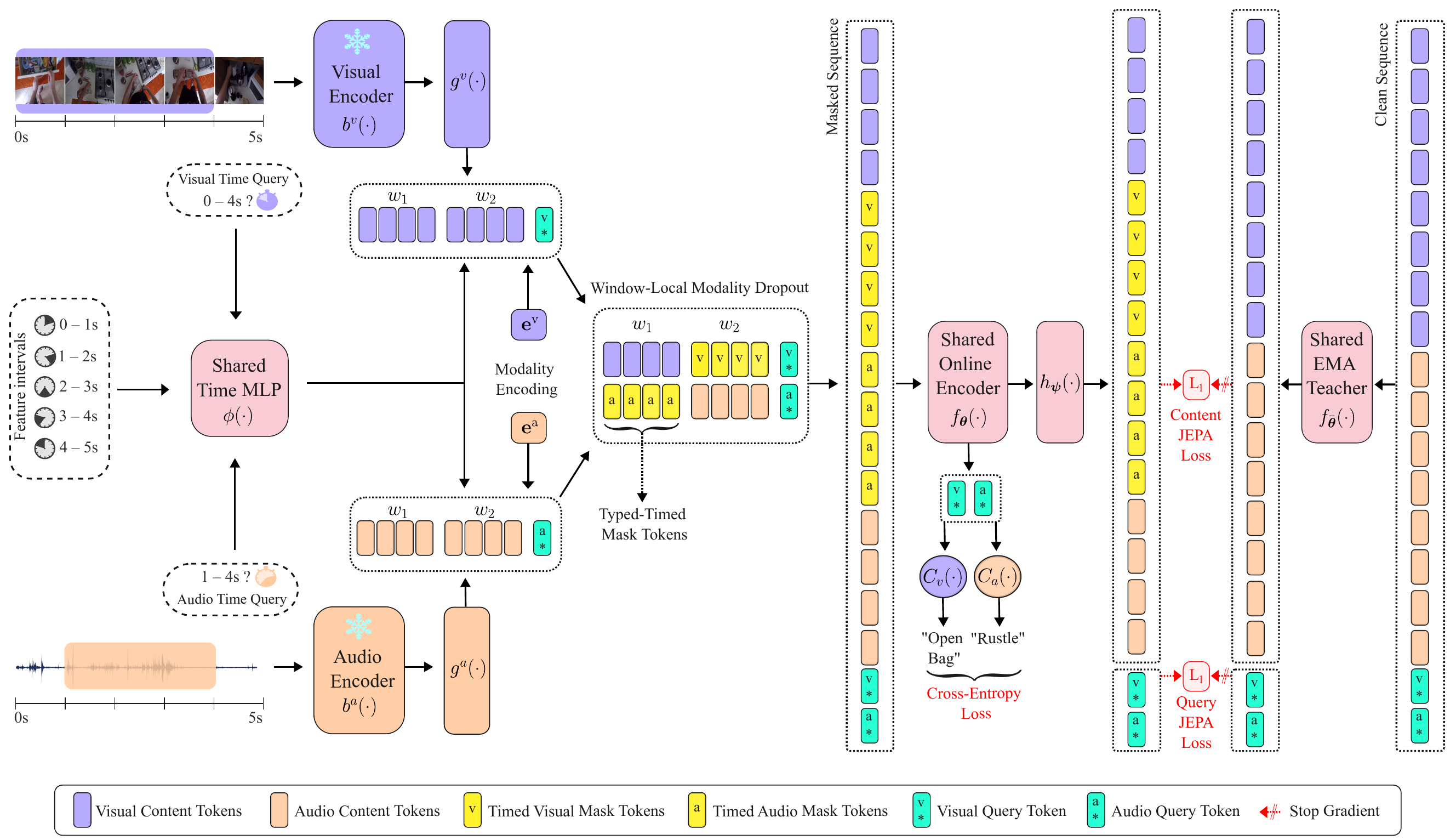}
\end{center}
\caption{\small \textbf{MacJEPA training framework.}
Window-local modality  dropout corrupts the online multimodal sequence with typed, timed masks, while content and query JEPA align it with clean EMA-teacher targets under joint task
supervision. Two windows and one query per stream are shown for clarity.}
\label{fig:main}
\end{figure}

\Cref{fig:main} summarizes MacJEPA. We first formulate interval-conditioned
recognition over untrimmed observations and construct its query sequence in
\Cref{sec:problem,sec:interval_queries}. We then introduce our two core
components: window-local modality dropout in \Cref{sec:modality_dropout} and
masked-to-clean content and query JEPA, trained jointly with recognition, in
\Cref{sec:query_jepa,sec:learning_objective}.

\subsection{Problem Setting}
\label{sec:problem}


We formulate audio-visual recognition directly on the untrimmed timeline rather
than around isolated event crops. For each modality
$m\in\Mset=\{\vmod,\amod\}$, a frozen unimodal encoder $\uenc{m}$ produces
temporally ordered features on a shared grid with stride $\fstride$\,seconds and
receptive field $\fextent=2\fstride$\,seconds. An observation is a contiguous
slice of $\Nrow$ rows,
$\featseq{m}=[\feat{m}{0},\ldots,\feat{m}{\Nrow-1}]$, spanning the nominal
duration $\obsspan=\Nrow\fstride$\,seconds. Candidate observations tile each video at
one-row intervals, independently of its annotations. They may therefore begin
or end within an event, include substantial background, or contain no annotated
visual action. An annotation is assigned to an observation when its intersection
with the observation is at least $\ovl=0.2$\,s, or when the annotation is fully
contained within it, following~\citet{chalk2024tim}. A training observation
contains multiple overlapping \EK visual action annotations and \ES  acoustic event annotations. The intervals
are supplied to the model rather than inferred, so the task is
interval-conditioned action recognition. To simulate temporal missingness, we
partition each observation into $\Nwin=\Nrow/\wsz$ non-overlapping windows of
$\wsz$ consecutive rows spanning $\windursym=\wsz\fstride$\,seconds.


\subsection{Interval Queries over an Untrimmed Observation}
\label{sec:interval_queries}

Multiple events share each observation, so time must identify both where evidence
occurs and which event is being predicted. For each modality $m\in\Mset$, a
modality-specific stem
$\stem{m}:\mathbb{R}^{\featdim{m}}\rightarrow\mathbb{R}^{\dcon}$ applies
feature dropout and projects the frozen features into a common content space. We
then encode every feature and query on the same temporal coordinate system. A
shared TimeMLP $\tmlp:\mathbb{R}^{2}\rightarrow\mathbb{R}^{\dtim}$ maps
continuous intervals to temporal representations, providing a universal clock
across both modalities~\citep{chalk2024tim}. Content row $j$ is assigned
$\rowiv{j}=[j\fstride,j\fstride+\fextent]/\obsspan$, while each annotation
interval is clipped to the observation and normalized to obtain
$\qiv{q}\in[0,1]^2$. A learned embedding
$\memb{m}\in\mathbb{R}^{\dmod}$ marks the modality of every token. Writing $j=k\wsz+f$, the content token $\ctok{m}{k}{f}$ at row $f$ of window $k$ and the token $\qtok{q}$ for
query $q$ are: 
\begin{equation}
\begin{aligned}
\ctok{m}{k}{f}
  &=
  \left[
    \stem{m}\!\left(\feat{m}{k\wsz+f}\right)
    \,\Vert\,
    \tmlp\!\left(\rowiv{k\wsz+f}\right)
  \right]+\memb{m},\\
\qtok{q}
  &=
  \left[
    \clsv{h(q)}
    \,\Vert\,
    \tmlp\!\left(\qiv{q}\right)
  \right]+\memb{m(q)} ,
\end{aligned}
\label{eq:token_construction}
\end{equation}
where $\clsv{h(q)}$ is a learned seed for the query head and $m(q)$ identifies
its stream. We concatenate $\dcon$ and $\dtim$, producing tokens of width
$\dmod=\dcon+\dtim$. The resulting sequence contains $\Ncon=2\Nrow$ multimodal content tokens, followed by verb, noun, and action queries for each \EK annotation and one sound
query for each \ES annotation. The encoder restricts content tokens to attending only to content, while each query attends to all content and itself but not other queries. This keeps the encoded context independent of the query set and each prediction independent of co-processed queries, with every classifier reading only its corresponding query representation. This construction temporally locates evidence and queries but assumes complete content; we model local absence next.

\subsection{Window-local Modality Dropout with Typed, Timed Masks}
\label{sec:modality_dropout}

We introduce window-local modality dropout to make modality availability vary within
the observation. The $\Nrow$ rows of an observation are partitioned into $\Nwin=\Nrow/\wsz$
non-overlapping windows of $\wsz$ rows, and a binary mask
$\Mask\in\{0,1\}^{\Nwin\times|\Mset|}$ specifies which modality is absent from
each window. In particular, $\Mel{k}{m}=1$ removes all $\wsz$ rows of modality
$m$ from window $k$; it is a window-level decision, not an independent
token mask. Windows are sampled independently, with probabilities
$1-\pdrop$, $\pdrop/2$, and $\pdrop/2$ for the complete, video-missing, and audio-missing states, respectively. Simultaneous loss of both streams is excluded so that every interval retains contemporaneous evidence from at least one sensor. Missing features are substituted and not deleted. Let
$\mskemb{m}\in\mathbb{R}^{\dcon}$ be a learned absence embedding for modality
$m$. The masked content token is: 
\begin{equation}
\widetilde{\vu}^{m}_{k,f}
=
\left[
  (1-\Mel{k}{m})
  \stem{m}\!\left(\feat{m}{k\wsz+f}\right)
  +\Mel{k}{m}\mskemb{m}
  \;\middle\Vert\;
  \tmlp\!\left(\rowiv{k\wsz+f}\right)
\right]
+\memb{m}.
\label{eq:mask_substitution}
\end{equation}
When $\Mel{k}{m}=0$, this recovers the observed content token from
\Cref{eq:token_construction}; when $\Mel{k}{m}=1$, only its projected content is replaced. The token remains in the sequence and retains its time encoding and modality tag. It therefore represents a typed, timed absence: the encoder is told which stream is unavailable and where. 
Furthermore, window grouping gives the simulated outage meaningful duration. Adjacent rows overlap by $50\%$, so masking one row would leave much of its temporal support available through its neighbors. Masking $\wsz$
rows creates a $\windursym=\wsz\fstride$ outage, which exceeds the
$\fextent=2\fstride$ receptive field of any single feature when $\wsz>2$. The windows are then flattened before encoding and impose no attention boundary. Every query still attends to all $\Ncon=2\Nrow$ content tokens, allowing it to route around a local outage and draw on evidence elsewhere in the observation. Conditional on missingness, the sampler has $3^{\Nwin}$ possible window configurations, redrawn each epoch, instead of the three states available to clip-level dropout; complete, video-missing, or audio-missing ~\citep{xiao2020avslowfast,ramazanova2025mmt,santosvillafranca2026karmma}. It therefore exposes a single event to partial evidence as sensors disappear and return. Whole-observation absence remains within its support as the limiting pattern $\Mel{k}{m}=1$ and $\Mel{k}{\bar m}=0$ for every $k$. Further details are in appendix \Cref{app:window-local-dropout}. We next regularize representations under these outages toward their clean counterparts.




\subsection{Content and Query JEPA under Simulated Outages}
\label{sec:query_jepa}

Let $\assemble(\featseq{\vmod},\featseq{\amod},\Qset;\Mask)$ denote the multimodal content-query sequence assembled under mask $\Mask$. A transformer encoder $\enc$ processes
the masked sequence, while an exponential-moving-average teacher $\tenc$
processes its clean counterpart (\Cref{fig:main}):
\begin{equation}
\begin{aligned}
(\Zctx_{\vtheta},\Zqry_{\vtheta})
  &=
  \enc\!\left(
    \assemble(\featseq{\vmod},\featseq{\amod},\Qset;\Mask)
  \right),\\
(\tZctx,\tZqry)
  &=
  \tenc\!\left(
    \assemble(\featseq{\vmod},\featseq{\amod},\Qset;\mathbf{0})
  \right).
\end{aligned}
\label{eq:jepa_branches}
\end{equation}
Both branches receive the same observations and queries; only the online branch
contains simulated outages. The teacher receives no gradients and is updated as
$\bar{\vtheta}\leftarrow
\emam\bar{\vtheta}+(1-\emam)\vtheta$, providing stable, contextualized clean
targets~\citep{grill2020byol,assran2023ijepa,baevski2022data2vec}. Predicting
these targets in a shared latent space avoids reconstructing heterogeneous audio
and visual signals in their incomparable input spaces~\citep{lei2025m3jepa}. 
A single token-wise predictor $\pred$, implemented as a
$\dmod\!\rightarrow\!384\!\rightarrow\!\dmod$ bottleneck MLP, maps online states ($\vz^{c}_{\vtheta}$, $\vz^{q}_{\vtheta}$)
towards their clean targets ($\bar{\vz}^{c}$, $\bar{\vz}^{q}$). It intentionally has no attention and receives no mask description
beyond the typed, timed absence already encoded by $\enc$. Predictor conditioning and capacity determine whether a JEPA
preserves or removes information about a transformation~\citep{garrido2024iwm}.
Our limited, unconditioned predictor favors outage-invariant readouts and
limits the train-only module from absorbing the cross-token fusion from the online encoder; more details are in appendix \Cref{app:predictor_EMA}.

For $\mathbf{u},\mathbf{v}\in\mathbb{R}^{\dmod}$, define
$d_1(\mathbf{u},\mathbf{v})=\|\mathbf{u}-\mathbf{v}\|_1/\dmod$. We regress both
content and valid query states $\Qvalid\subseteq\Qset$ onto parameter-free normalized, stop-gradient
teacher targets:
\begin{equation}
\begin{aligned}
\lctx
  &=
  \frac{1}{\Ncon}
  \sum_{i=1}^{\Ncon}
  d_1\!\left(
    \pred(\vz^{c}_{\vtheta,i}),
    \lnz{\sg{\bar{\vz}^{c}_{i}}}
  \right),\\
\lqry
  &=
  \frac{1}{|\Qvalid|}
  \sum_{q\in\Qvalid}
  d_1\!\left(
    \pred(\vz^{q}_{\vtheta,q}),
    \lnz{\sg{\bar{\vz}^{q}_{q}}}
  \right),\\
\Ljepa
  &= \lctx+\lqry .
\end{aligned}
\label{eq:context_query_jepa}
\end{equation}
The two terms are normalized separately, giving content and query states equal
total weight. Unlike masked-position objectives
~\citep{baevski2022data2vec,assran2023ijepa,teotia2026mjepa}, $\lctx$ covers all
$\Ncon=2\Nrow$ content states. Since global attention propagates corruption to observed tokens, this aligns
their context-dependent representations with clean targets rather than merely
filling masked positions. Crucially, content states cannot attend to queries, so
$\partial\lctx/\partial\clsv{h}=0$. Although $\lctx$ updates the shared encoder, it cannot directly regularize the learned query seed read by classifier $C_h$. The query term instead aligns the same task and interval under corrupted and clean context. It therefore places masked-to-clean consistency directly on the prediction representation, subsequently supervised by cross-entropy. This encourages missing-modality invariance in task-conditioned readouts. Window-local dropout creates outage conditions to enable query attention to reroute toward surviving evidence. Content JEPA stabilizes content representations, and query JEPA preserves the prediction readout, allowing robustness to emerge as the encoder combines surviving temporal and cross-modal evidence in latent space. The teacher and predictor are discarded after training.


\subsection{Learning Objective}
\label{sec:learning_objective}

For task heads
$\Hset=\{\mathrm{verb},\mathrm{noun},\mathrm{action},\mathrm{sound}\}$, let
$\Qhead{h}\subseteq\Qvalid$ denote the valid, unpadded queries for head $h$. Each classifier
$C_h$ predicts from its corresponding interval-conditioned query readouts using
label-smoothed cross-entropy with $\varepsilon=0.1$:
\begin{equation}
\Lce
=
\sum_{h\in\Hset}
\frac{1}{|\Qhead{h}|}
\sum_{q\in\Qhead{h}}
\operatorname{CE}_{\varepsilon}\!\left(
  C_h(\vz^{q}_{\vtheta,q}),
  \qlabel{q}
\right).
\label{eq:classification_objective}
\end{equation}
We additionally apply VICReg to observed content states as a representational collapse guard,
excluding substituted tokens so absence embeddings do not match real sensor
statistics~\citep{bardes2022vicreg}. All objectives are optimized jointly; complete details are in appendix \Cref{app:complete-objective,app:training-schedule}:

\begin{equation}
\Ltot
=
w_{\mathrm{ce}} \Lce
+
w_{\mathrm{jepa}} \Ljepa
+
w_{\mathrm{vic}} \Lvic .
\label{eq:total_objective}
\end{equation}

%% file: sections/04_experiments.tex
\section{Experiments}
\label{sec:experiments}


For all experiments, we set $\fstride=16/30\,\mathrm{s}$, $\fextent=32/30\,\mathrm{s}$, $\Nrow=36$, $\wsz=4$, $\dcon=768$, $\dtim=256$, $\pdrop=0.7$. This gives $\obsspan=19.2\,\mathrm{s}$, $\Nwin=9$, $\windursym=2.13\,\mathrm{s}$, $\Ncon=72$, and $\dmod=1024$.
Following the setup in \Cref{sec:experimental_setup}, we examine how MacJEPA
learns missing-modality robustness through controlled design hypotheses
(\Cref{subsec:design-hypotheses}) and qualitative evidence-routing analysis
(\Cref{qual-analysis}), then compare the same fixed checkpoint with published
benchmarks under complete input and across the missingness spectrum
(\Cref{sec:prior-comparison}).


\subsection{Experimental Setup}
\label{sec:experimental_setup}

\textbf{Data.}
We jointly evaluate visual actions from \EK-100 and acoustic events from \ES, which
annotate the same 100 hours of untrimmed egocentric video with distinct temporal
intervals~\citep{damen2022rescaling,huh2023epicsounds}. \EK contains 97 verbs
and 300 nouns; we classify the 3,806 verb--noun pairs or action labels present in the training and validation sets, while
\ES contains 44 sound classes. We use the official training and validation
annotations. We evaluate 9,668 \EK validation and 8,045 \ES validation
segments.

\textbf{Setup.}
We concatenate frozen Omnivore Swin-B and VideoMAE-L features into a
$2{,}048$-D visual input and use $2{,}304$-D Auditory SlowFast R50 audio
features~\citep{girdhar2022omnivore,tong2022videomae,kazakos2021asf}. Each
19.2\,s observation has 36 aligned rows grouped into nine 2.13\,s windows.
768-D content and 256-D time representations form 1,024-D tokens for a
6-layer, 8-head transformer with MLP ratio 2. Training uses AdamW for 50
epochs with batch size 256, learning rate $3\times10^{-4}$, weight decay
$10^{-4}$, two-epoch warm-up, and cosine decay. During training, each observation window independently drops one uniformly chosen modality with probability 0.7. 
Evaluation uses two missingness protocols. \textbf{Nested per-window temporal
missingness.} To evaluate robustness to local sensor disappearance and recovery within an observation, controlled analyses in \Cref{subsec:design-hypotheses} and appendix \Cref{app:exp_ex} use fixed nested masks
that remove the selected modality from each observation window with probability $r\in\{0,0.25,0.50,0.75,1\}$. At $r=0.75$, this affects 6.75 of nine windows
on average, equivalent to 27 aligned rows or 14.4\,s of the validation observations. \textbf{Observation-level
missingness.} For comparison with prior work in
\Cref{sec:exp:robust}, we follow \textsc{MMT} and \textsc{MiDl} to define missingness; a fraction $r$
of validation observations loses the selected modality for the full $19.2$\,s,
while the remainder stay complete
~\citep{ramazanova2025mmt,ramazanova2025midl}.
Following~\citet{chalk2024tim}, we average logits across overlapping observations containing each annotation and report annotation-level top-1 accuracy. Video is dominant for \EK and audio for \ES, with the reverse defining weaker-modality removal. All results use the same epoch-50 checkpoint. Appendix \Cref{app:implementation,app:exp_ex} details feature extraction, implementation,
optimization, and ablations of drop probabilities, window
and observation lengths, feature backbones, and encoder depth.


\subsection{What Produces Missing-Modality Robustness in MacJEPA?}
\label{subsec:design-hypotheses}

\begin{table}[t]
\caption{\small Controlled component analysis under dominant-modality removal. Top-1 accuracy (\%) for \EK action and
\ES sound reported. Follows nested per-window temporal missingness protocol.}
\label{tab:component-analysis}
\centering
\footnotesize
\setlength{\tabcolsep}{2pt}
\setlength{\arrayrulewidth}{0.7pt}
\renewcommand{\arraystretch}{1.2}

\begin{tabularx}{\linewidth}{
  @{}>{\raggedright\arraybackslash}X
  *{5}{>{\centering\arraybackslash}p{0.042\linewidth}} |
  *{5}{>{\centering\arraybackslash}p{0.042\linewidth}}@{}
}
\hline
Ablation &
\multicolumn{5}{c|}{\EK action, V drop} &
\multicolumn{5}{c}{\ES sound, A drop} \\
\cline{2-6}
\cline{7-11}
& 0 & 25 & 50 & 75 & 100
& 0 & 25 & 50 & 75 & 100 \\
\hline
\emph{Unimodal reference}
& 13.0 & 13.0 & 13.0 & 13.0 & 13.0
& 41.4 & 41.4 & 41.4 & 41.4 & 41.4 \\
\hline
H1: Encoder + CE
& 53.6 & 51.6 & 45.7 & 27.8 & 4.1
& 56.5 & 55.1 & 53.5 & 45.2 & 26.3 \\
H2: H1 + window-local modality dropout
& 52.9 & 51.7 & 48.7 & 44.0 & 16.3
& 55.5 & 54.4 & 53.8 & 51.0 & 46.7 \\
H3: H2 + content JEPA
& 53.6 & 53.1 & 50.8 & 45.2 & 18.2
& 56.1 & 55.5 & \textbf{55.0} & 51.5 & 48.4 \\
\hline
H4: H3 + query JEPA (\textbf{MacJEPA})
& \textbf{54.0} & \textbf{53.7} & \textbf{52.2}
& \textbf{47.3} & \textbf{19.5}
& \textbf{58.1} & \textbf{56.0} & 54.3
& \textbf{52.1} & \textbf{50.0} \\
\hline
\end{tabularx}
\end{table}

\Cref{tab:component-analysis} isolates each component's robustness contribution under
dominant-modality removal following the nested per-window temporal missigness defined in \Cref{sec:experimental_setup}, fixing the features, encoder, query interface,
and training schedule. Following prior work, the unimodal references use only
the surviving weaker stream, testing whether robust multimodal inference remains
preferable~\citep{ramazanova2025mmt,ramazanova2025midl}. On \EK, verbs, e.g., \textit{Open}, describe temporally extended interactions, while nouns, e.g., \textit{Bag}, identify manipulated objects from fewer salient frames, and their composition defines actions, e.g., \textit{Open Bag}. On \ES, the sound head identifies acoustic events, e.g., \textit{Rustle}.
\textbf{(1) Missingness training is necessary.}
H1 sees only complete observations, leaving its absence embeddings untrained.
Under dominant-stream loss, audio may disclose an \EK verb but rarely its
visually grounded noun. Similarly, recovering an \ES sound from vision
requires learning that correspondence. H1 therefore falls from 53.6 to 4.1 on
\EK and from 56.5 to 26.3 on \ES, respectively 8.9 and 15.1 points below the
unimodal references, confirming the fragility of conventional multimodal models
~\citep{ramazanova2025mmt,ramazanova2025midl}
\textbf{(2) Temporal outage exposure improves robustness.}
Window-local modality dropout with learned mask tokens makes absence temporally
addressable. Since verb cues unfold across an interval, masking encourages H2 to integrate surviving windows rather than depend on one frame. Nouns often rely on fewer salient object-defining frames, while the sound head loses roughly one-third of its acoustic evidence, making clean supervision for them less reliable. H2 consequently raises the $100\%$ endpoints by 12.2 and 20.4 points but lowers clean action and sound accuracy by 0.7 and 1.0 points, reflecting the robustness--clean-accuracy trade-off
~\citep{ramazanova2025mmt}.
\textbf{(3) Content JEPA restores contextual consistency.}
Since global attention propagates an outage beyond the replaced tokens, dense
content JEPA aligns all corrupted content representations with clean
targets, stabilizing the multimodal context. For action recognition, it restores the noun recognition object cues weakened by dropout, without discarding the temporal integration learned for verbs, although \Cref{tab:component-analysis} does not isolate noun and verb effects separately.  Consequently, H3 recovers H1's clean \EK action ceiling and comes within
0.4 points on \ES sound, while further improving on the $100\%$ endpoints of H2 by 1.9
and 1.7 points.
\textbf{(4) Query JEPA aligns what classifiers read.}
Content JEPA cannot directly constrain interval-conditioned queries,
since content tokens never attend to them. Query JEPA aligns the masked
and clean classifier readouts, preserving the verb--noun composition of visual action and the cross-modal evidence supporting sound. H4 improves over H3 by 0.4 and 2.0 points under complete input and by 1.3 and 1.6 points under total dominant-stream loss. Together, dropout simulates the condition for
attention to reroute toward surviving evidence, content JEPA stabilizes the
multimodal context, and query JEPA preserves the task decision.

\subsection{MacJEPA Queries Reroute Attention Around Local Outages}
\label{qual-analysis}

\begin{figure}[t]
\begin{center}
\includegraphics[width=\linewidth, height=7.6cm]{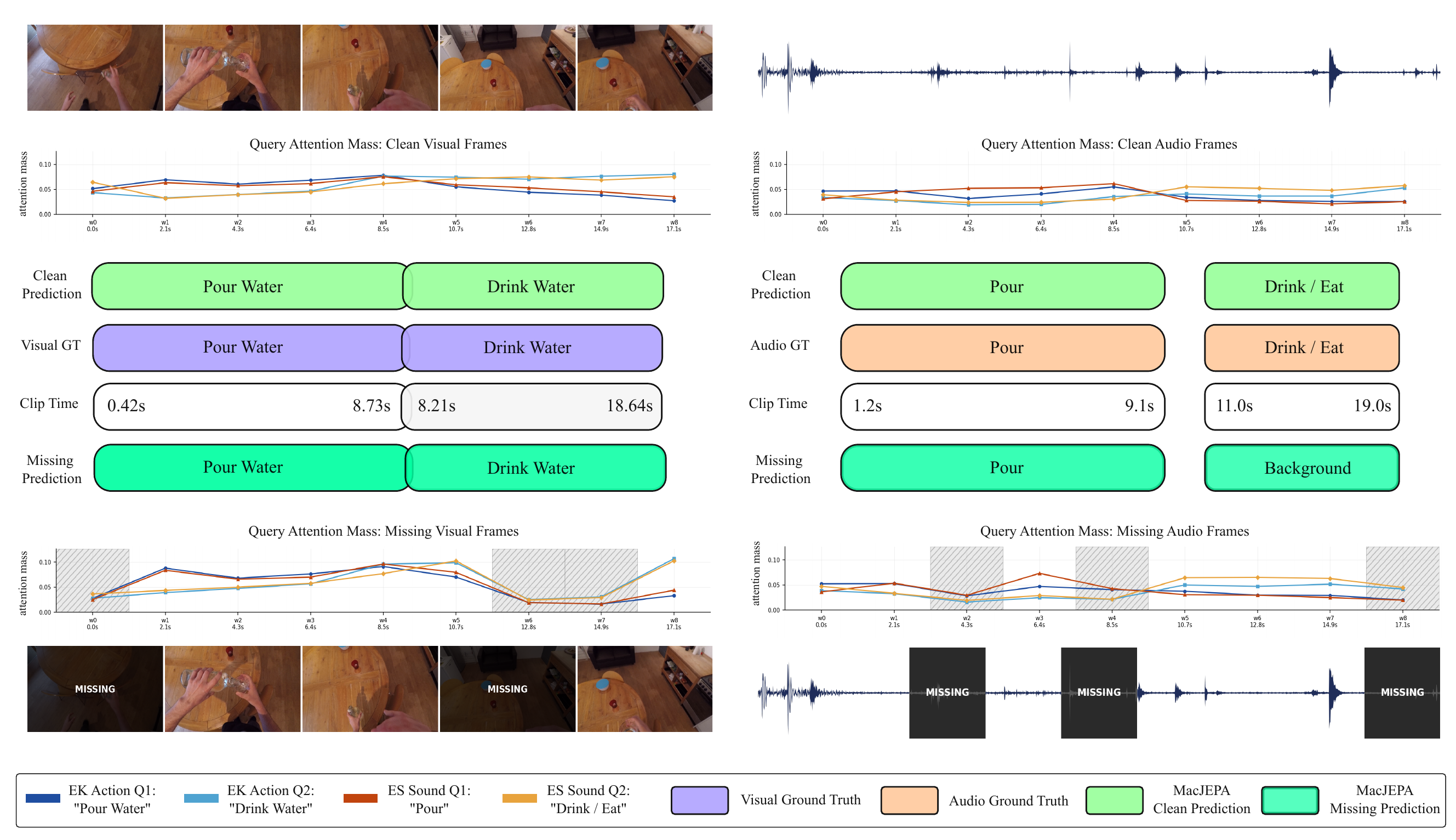}
\end{center}
\caption{ \small Query attention over visual (left) and audio (right) windows
before (above) and after (below) local outages. Hatched regions are missing, boxes compare
clean and corrupted predictions against ground truth.}
\label{fig:qual}
\end{figure}

\Cref{fig:qual} follows four interval queries within one 19.2\,s validation observation;
each curve shows the fraction of a query's attention assigned to each 2.13\,s
visual or audio interval, averaged across encoder layers and heads. With complete
input, the queries attend to distinct temporal regions, and all four predictions
match their ground truth. The corruption removes video over 0--2.1\,s and
12.8--17.1\,s, and audio over 4.3--6.4\,s, 8.5--10.7\,s, and
17.1--19.2\,s, disrupting both the early \emph{Pour} and later \emph{Drink}
queries. Attention responds locally: the early queries shift visual mass from
0--2.1\,s to 2.1--4.3\,s, while the \emph{Pour} sound query moves its audio
peak from 8.5--10.7\,s to the surviving 6.4--8.5\,s interval. The later
queries redistribute visual mass from 12.8--17.1\,s to their neighboring
intervals, while their audio peaks move from the missing final interval to
14.9--17.1\,s for \emph{Drink Water} and 12.8--14.9\,s for
\emph{Drink/Eat}. Under this attention redistribution, both \EK predictions and the \ES \emph{Pour} prediction remain unchanged; only \emph{Drink/Eat} changes to \emph{Background} since substantial late multimodal evidence is removed. In this example, we visualize how query attention mass redistributes away from local outages toward nearby surviving evidence, while three of the four task readouts remain unchanged relative to clean input.



\subsection{Comparison with Prior Work}
\label{sec:prior-comparison}

\subsubsection{Full-modality Recognition}
\label{sec:full-modality}

\begin{table}[t]
\caption{\small Full-modality validation top-1 accuracy (\%). MacJEPA uses a single
saved checkpoint for both \EK and \ES. Best results are bold, and second-best results are underlined.}
\label{tab:full-modality}
\centering
\small
\setlength{\tabcolsep}{2.5pt}
\renewcommand{\arraystretch}{1}

\begin{minipage}[t]{0.62\linewidth}
\vspace{0pt}
\centering
(a) \EK\\[0.2em]
\begin{tabularx}{\linewidth}{
  @{}>{\raggedright\arraybackslash}X c c c c@{}
}
\toprule
Audio-Visual Benchmarks & \textit{xp} &
Verb & Noun & Action \\
\midrule
\textsc{TBN}
  & 224p & 66.0 & 47.2 & 36.7 \\
\textsc{MMT}
  & 224p & 64.0 & 57.3 & 42.8 \\
\textsc{MBT}
  & 224p & 64.8 & 58.0 & 43.4 \\
\textsc{MTCN}
  & 336p & 70.7 & 62.1 & 49.6 \\
\textsc{M\&M}
  & 420p & 72.0 & \underline{66.3} & 53.6 \\
\textsc{TIM}
  & 224p & \textbf{77.1} & \textbf{67.2} & \textbf{57.5} \\
\midrule
\textbf{MacJEPA (ours)}
  & 224p & \underline{74.0} & 64.5 & \underline{54.0} \\
\bottomrule
\end{tabularx}
\end{minipage}
\hfill
\begin{minipage}[t]{0.35\linewidth}
\vspace{0pt}
\centering
(b) \ES\\[0.2em]
\begin{tabularx}{\linewidth}{
  @{}>{\raggedright\arraybackslash}X c c@{}
}
\toprule
Benchmarks & Mod. & Sound \\
\midrule
\textsc{SSAST}
  & A & 53.5 \\
\textsc{ASF}
  & A & 53.8 \\
\textsc{MMT}
  & A,V & 55.2 \\
\textsc{TIM}
  & A & 55.7 \\
\textsc{DiffSED}
  & A & 56.9 \\
\textsc{TIM}
  & A,V & \textbf{58.3} \\
\midrule
\textbf{MacJEPA (ours)}
  & A,V & \underline{58.1} \\
\bottomrule
\end{tabularx}
\end{minipage}

\vspace{0.35em}
\end{table}

\Cref{tab:full-modality} establishes two findings under complete input.
\textbf{(1.)} On \EK, \Cref{tab:full-modality}(a), MacJEPA achieves 74.0 verb, 64.5 noun, and 54.0 action accuracy
using 224p inputs, ranking second on verb and action while outperforming
\textsc{TBN}, \textsc{MBT}, \textsc{MTCN}, and \textsc{M\&M} on both metrics
despite the latter using 420p inputs
~\citep{kazakos2019epicfusion,nagrani2021mbt,kazakos2021mtcn,
xiong2022mmmix,chalk2024tim}. The \textsc{MMT} row, the only
missing-modality framework in this comparison, reports its modal-complete
\textsc{MBT} baseline rather than its adapted model
~\citep{ramazanova2025mmt}. \textbf{(2.)} On \ES, \Cref{tab:full-modality}(b), MacJEPA reaches 58.1 sound accuracy,
only 0.2 points below audio-visual \textsc{TIM}, 1.2 points above the strongest
audio-only model, \textsc{DiffSED}, and 2.9 points above the modal-complete
\textsc{MMT} baseline
~\citep{chalk2024tim,bhosale2024diffsed,ramazanova2025mmt}. It also surpasses
\textsc{SSAST}, \textsc{ASF}, and audio-only \textsc{TIM}
~\citep{gong2022ssast,kazakos2021asf,chalk2024tim}. 

\begin{table}[t]
\caption{\small Dominant-modality robustness. Top-1 accuracy (\%) for \EK verb under
missing video and \ES sound under missing audio. Follows observation-level missingness protocol.}
\label{tab:dominant-missing}
\centering
\small
\setlength{\tabcolsep}{2.5pt}
\renewcommand{\arraystretch}{1}

\begin{minipage}[t]{0.495\linewidth}
\vspace{0pt}
\centering
(a) \EK verb: Video dropped\\[0.2em]
\begin{tabularx}{\linewidth}{
  @{}>{\raggedright\arraybackslash}X c c c c c@{}
}
\toprule
Benchmarks &
\multicolumn{5}{c}{Missing rate (\%)} \\
\cmidrule(lr){2-6}
& 0 & 25 & 50 &
  75 & 100 \\
\midrule
\emph{Unimodal (A only)}
  & 40.0 & 40.0 & 40.0 & 40.0 & 40.0 \\
\textsc{MiDl-LTA}
  & 63.7 & 58.4 & 52.4 & 46.7 & 41.4 \\
\textsc{MMT}
  & 63.4 & 59.0 & 54.6
  & 50.0 & 45.3 \\
\midrule
\textbf{MacJEPA (ours)}
  & \textbf{74.0} & \textbf{67.8} & \textbf{61.6}
  & \textbf{55.6} & \textbf{49.5} \\
\bottomrule
\end{tabularx}
\end{minipage}
\hfill
\begin{minipage}[t]{0.495\linewidth}
\vspace{0pt}
\centering
(b) \ES sound: Audio dropped\\[0.2em]
\begin{tabularx}{\linewidth}{
  @{}>{\raggedright\arraybackslash}X c c c c c@{}
}
\toprule
Benchmarks &
\multicolumn{5}{c}{Missing rate (\%)} \\
\cmidrule(lr){2-6}
& 0 & 25 & 50 &
  75 & 100 \\
\midrule
\emph{Unimodal (V only)}
  & 41.4 & 41.4 & 41.4 & 41.4 & 41.4 \\
\textsc{MiDl-LTA}
  & 54.9 & 46.8 & 39.5 & 32.6 & 26.0 \\
\textsc{MMT}
  & 56.3 & 52.3 & 48.6
  & 44.6 & 40.7 \\
\midrule
\textbf{MacJEPA (ours)}
  & \textbf{58.1} & \textbf{55.9} & \textbf{53.9}
  & \textbf{51.8} & \textbf{50.0} \\
\bottomrule
\end{tabularx}
\end{minipage}
\end{table}

\begin{table}[t]
\caption{\small Auxiliary-modality robustness. Top-1 accuracy (\%) for \EK verb under
missing audio and \ES sound under missing video. Follows observation-level missingness protocol.}
\label{tab:weak-missing}
\centering
\small
\setlength{\tabcolsep}{2.5pt}
\renewcommand{\arraystretch}{1}

\begin{minipage}[t]{0.495\linewidth}
\vspace{0pt}
\centering
(a) \EK verb: Audio dropped\\[0.2em]
\begin{tabularx}{\linewidth}{
  @{}>{\raggedright\arraybackslash}X c c c c c@{}
}
\toprule
Benchmarks &
\multicolumn{5}{c}{Missing rate (\%)} \\
\cmidrule(lr){2-6}
& 0 & 25 & 50 &
  75 & 100 \\
\midrule
\emph{Unimodal (V only)}
  & 63.2 & 63.2 & 63.2 & 63.2 & 63.2 \\
\textsc{MiDl-LTA}
  & 63.7 & 61.7 & 59.5
  & 57.3 & 55.3 \\
\midrule
\textbf{MacJEPA (ours)}
  & \textbf{74.0} & \textbf{73.5} & \textbf{73.2}
  & \textbf{72.9} & \textbf{72.5} \\
\bottomrule
\end{tabularx}
\end{minipage}
\hfill
\begin{minipage}[t]{0.495\linewidth}
\vspace{0pt}
\centering
(b) \ES sound: Video dropped\\[0.2em]
\begin{tabularx}{\linewidth}{
  @{}>{\raggedright\arraybackslash}X c c c c c@{}
}
\toprule
Benchmarks &
\multicolumn{5}{c}{Missing rate (\%)} \\
\cmidrule(lr){2-6}
& 0 & 25 & 50 &
  75 & 100 \\
\midrule
\emph{Unimodal (A only)}
  & 46.5 & 46.5 & 46.5 & 46.5 & 46.5 \\
\textsc{MiDl-LTA}
  & 55.0 & 53.4 & 52.0
  & 51.0 &49.4 \\
\midrule
\textbf{MacJEPA (ours)}
  & \textbf{58.1} & \textbf{56.9} & \textbf{55.6}
  & \textbf{54.4} & \textbf{53.2} \\
\bottomrule
\end{tabularx}
\end{minipage}
\end{table}

\subsubsection{Missing-Modality Robustness}
\label{sec:exp:robust}


\Cref{tab:dominant-missing,tab:weak-missing} adopt the observation-level missingness of \Cref{sec:experimental_setup} to align with published benchmarks, which score trimmed single-event clips using separate \EK and \ES models. Our MacJEPA model is instead jointly evaluated on both annotation sets within each 19.2\,s
untrimmed validation video, containing on average 4.55 \EK and 4.04 \ES annotations.
\textbf{Robustness to Dominant Missing Modalities.} Following prior work, the unimodal references are trained only on the auxiliary surviving
stream~\citep{ramazanova2025mmt,ramazanova2025midl}.
\Cref{tab:dominant-missing} reveals three findings.
\textbf{(1)} MacJEPA leads both benchmarks at every missing rate, indicating that its queries
can increasingly rely on surviving evidence even at this observation-level missingness protocol.
\textbf{(2)} At $75\%$ missingness, it exceeds
\textsc{MMT} by 5.6 and 7.2 points; on \EK and \ES respectively, where \textsc{MiDl-LTA} has already fallen below the
41.4 unimodal reference on \ES. \textbf{(3)} At the $100\%$ endpoint, MacJEPA leads the strongest prior method by 4.2 points on \EK and 9.3 points on \ES. MacJEPA therefore achieves the highest reported absolute accuracy across the dominant-modality missingness sweep, including complete stream loss.
\textbf{Robustness to Auxiliary Missing Modalities.} Following prior work, our unimodal references use only the dominant surviving stream.~\citep{nagrani2021mbt,ramazanova2025mmt}. \Cref{tab:weak-missing} yields two findings.
\textbf{(1)} MacJEPA loses only 1.5 points when audio is removed from all observations on \EK and 4.9
points when all video is removed on \ES, compared with drops of 8.4 and 5.6
points for \textsc{MiDl-LTA}, respectively
~\citep{ramazanova2025midl}. 
\textbf{(2)} At $100\%$ missingness, MacJEPA exceeds \textsc{MiDl-LTA} by 17.2 points
on \EK and 3.8 points on \ES, while remaining 9.3 and 6.7 points above the
corresponding dominant-stream unimodal references.  
Supported by our component and attention analyses, a single fixed MacJEPA
checkpoint remains competitive with established audio-visual models under
complete input, withstands local outages under our nested per-window temporal missingness protocol,
and outperforms prior missing-modality methods under their observation-level missingness protocol across dominant and auxiliary removal, including complete stream loss, without test-time adaptation.




%% file: sections/05_conclusion.tex
\section{Conclusion}
\label{sec:conclusion}
MacJEPA redefines egocentric missingness as local sensor outages within
untrimmed, multi-event observations, with whole-clip absence as the limiting
case, and addresses it by introducing typed interval-wise dropout and supervised
content-query JEPA. A single checkpoint remains competitive with complete input
and leads prior methods at every reported dominant- and auxiliary-stream missing
rate on \EK and \ES. Future work can extend this framework to temporal action
detection under missingness and additional egocentric streams such as optical
flow and IMU.

%% file: sections/07_statements.tex

\subsection*{AI use statement}
In this work, generative AI tools were used only to polish the manuscript's language, including grammar correction, sentence rephrasing, and improvements to readability and clarity. They were not used to generate data, develop theoretical models or conceptual frameworks, formulate mathematical claims, assist with proofs, propose or refine hypotheses, design or evaluate the methodology or experiments, implement methods, translate, clean or reformat data, conduct qualitative or thematic analysis, or interpret results. The authors developed all research concepts and ideas and conducted the analyses.  The authors reviewed all AI-assisted text for accuracy, originality, and compliance with ethical standards, and take full responsibility for the final manuscript and all of its claims.

\subsection*{Ethics statement}
This work adheres to the ICLR Code of Ethics. No new human-subject or animal
experimentation was conducted. All datasets were obtained and used in accordance
with their respective usage guidelines. We did not seek to identify participants
or collect or release additional personal information. We considered potential
privacy, bias, and security concerns and remain committed to transparent and
responsible reporting throughout the research process.

\subsection*{Reproducibility statement}

We provide a code repository containing the MacJEPA scripts, implementation, and
reproduction instructions in a zipped code file in our supplementary materials.
\Cref{sec:experimental_setup} specifies the experimental and evaluation
protocols, while \Cref{app:implementation} details feature extraction, model
construction, optimization, and implementation choices. Together, these
resources document the complete pipeline required to reproduce our results.

%
%
\subsubsection*{Acknowledgments}
This work was supported by the Federal Ministry of Research, Technology and Space of Germany [project name: EMuLE – Enhancing Data and Model Efficiency in Multimodal Learning; grant number 16IS24059]. The last author was partially funded by the Lower Saxony Ministry of Science and Culture (MWK) with funds from the Volkswagen Foundation’s Zukunft Niedersachsen program [project name: CAIMed - Lower Saxony Center for Artificial Intelligence and Causal Methods in Medicine; grant number: ZN4257]. The authors acknowledge the Hannover Medical School for providing MHH-HPC resources that have contributed to the results reported in this paper.

%% file: sections/06_appendix.tex
\section*{Appendix}

\section{MacJEPA Framework: Extended Discussion}
\label{app: discussion}

The interval coordinates, token construction, encoder architecture, attention
mask, and task heads follow the formulation in
\Cref{sec:method} and are specified fully in
\Cref{app:implementation}. In brief, each observation is represented by  $\Ncon=2\Nrow$
time-encoded audio-visual content tokens followed by interval-conditioned verb,
noun, action, and sound queries. Content tokens attend globally to content,
while each query attends to all content and itself but not to other queries.
In this section, we primarily focus on MacJEPA's two central mechanisms: representing temporally local sensor outages and aligning their resulting content and task readouts
with clean latent targets. Additionally, we detail the task-specific prediction heads we used, the VICReg loss and the complete learning objective of our framework in this section.

\subsection{Window-Local Modality Dropout and Typed, Timed Masking}
\label{app:window-local-dropout}

MacJEPA represents availability with a binary mask
$\Mask\in\{0,1\}^{\Nwin\times|\Mset|}$, where $\Mel{k}{m}=1$ means that
modality $m$ is unavailable throughout window $k$. When window-local
corruption is applied, each window independently draws one of three states:
\begin{equation}
s_k
\sim
\operatorname{Categorical}
\left(
1-\pdrop,\,
\frac{\pdrop}{2},\,
\frac{\pdrop}{2}
\right),
\qquad
s_k\in\{0,\vmod,\amod\},
\label{eq:app-window-sampler}
\end{equation}
with
\begin{equation}
\Mel{k}{m}=\mathbf{1}[s_k=m].
\label{eq:app-window-mask}
\end{equation}
The state $s_k=0$ preserves both streams, while $s_k=\vmod$ and $s_k=\amod$
remove vision and audio, respectively. Removing both is excluded, ensuring that
every window retains contemporaneous evidence from at least one sensor. For
either modality,
\begin{equation}
\Pr\!\left(\Mel{k}{m}=1\right)=\frac{\pdrop}{2}.
\label{eq:app-per-modality-drop}
\end{equation}
Since each window contributes equal numbers of visual and audio tokens,
$\pdrop/2$ is also the expected fraction of substituted tokens in the complete
multimodal sequence. Across $\Nwin$ windows, the sampler supports
\begin{equation}
3^{\Nwin}
\label{eq:app-corruption-support}
\end{equation}
patterns of disappearance and recovery, redrawn each epoch. These patterns are
not equiprobable because the complete, video-missing, and audio-missing states
occur with probabilities $1-\pdrop$, $\pdrop/2$, and $\pdrop/2$,
respectively. Whole-observation absence remains within this support when one
modality is removed from every window, recovering clip-level missingness as a
limiting case.

Let $\mskemb{m}\in\mathbb{R}^{\dcon}$ be the learned absence embedding for
modality $m$. The masked content token is
\begin{equation}
\widetilde{\vu}^{m}_{k,f}
=
\left[
(1-\Mel{k}{m})
\stem{m}\!\left(\feat{m}{k\wsz+f}\right)
+
\Mel{k}{m}\mskemb{m}
\;\middle\Vert\;
\tmlp\!\left(\rowiv{k\wsz+f}\right)
\right]
+\memb{m}.
\label{eq:app-mask-substitution}
\end{equation}
The projected feature is replaced, but its sequence position is not deleted.
The token retains its temporal encoding and modality tag. The learned vector
$\mskemb{m}$ identifies which stream is missing, while the complete token
$\widetilde{\vu}^{m}_{k,f}$ additionally identifies where that absence occurs.
Visual and audio use separate absence embeddings. Since corresponding rows share
identical temporal coordinates, a common absence vector would give both streams
the same substituted content component. Modality-specific vectors make the
missing stream explicit before attention is applied.

Windowing also gives the corruption meaningful duration. We choose $\wsz$ such
that
\begin{equation}
\windursym=\wsz\fstride>\fextent .
\label{eq:app-window-duration}
\end{equation}
Because adjacent feature supports overlap, removing one row would leave much of
its temporal evidence available through neighboring features. Masking $\wsz$
consecutive rows instead creates an outage whose duration exceeds the receptive
field of an individual feature and cannot be recovered from feature overlap
alone.

Importantly, substitution does not explicitly block attention to missing
positions. For query $q$ and content token $i$, one attention head computes
\begin{equation}
\alpha_{qi}
=
\operatorname{softmax}_{i}\!\left(
\frac{
(W_Q\vz^{q}_{q})^{\top}
(W_K\widetilde{\vu}^{c}_{i})
}{
\sqrt{d_h}
}
\right),
\qquad
\mathbf{o}_{q}
=
\sum_{i=1}^{\Ncon}
\alpha_{qi}W_V\widetilde{\vu}^{c}_{i},
\label{eq:app-query-rerouting}
\end{equation}
where $W_Q$, $W_K$, and $W_V$ are the attention projections and $d_h$ is the
head width. A typed, timed absence token remains visible as a key and value,
allowing the query to recognize that a particular source is unavailable at that
time. Masked-to-clean query alignment then encourages the query to reduce its
dependence on unavailable evidence and redistribute attention toward the
surviving modality or neighboring temporal context. This rerouting is learned
from the robustness objective rather than imposed by an attention mask.
The same substitution also changes content self-attention. Every observed
content token can attend to absence tokens and surviving evidence across the
observation. Dense content JEPA constrains these context-dependent content
states toward their clean targets, while query JEPA refines the resulting
task-specific attention readout. Mask tokens therefore serve two connected
roles: they stabilize content reasoning around an outage and, more directly,
enable each query to reroute toward evidence that remains informative.
Finally, windows define only the support of corruption. They are flattened
before encoding and impose no attention boundaries. Every query retains access
to all $\Ncon=2\Nrow$ content tokens across the complete
$\obsspan=\Nrow\fstride$ observation.

\subsection{EMA Teacher, Predictor, and JEPA}
\label{app:predictor_EMA}

For observation
$\mathbf{X}=(\featseq{\vmod},\featseq{\amod})$, query set $\Qset$, and sampled
mask $\Mask$, the online and target branches compute
\begin{equation}
\begin{aligned}
(\Zctx_{\vtheta},\Zqry_{\vtheta})
&=
\enc\!\left(
\assemble(\mathbf{X},\Qset;\Mask)
\right),\\
(\bar{\mathbf{Z}}^{c},\bar{\mathbf{Z}}^{q})
&=
\tenc\!\left(
\assemble(\mathbf{X},\Qset;\mathbf{0})
\right).
\end{aligned}
\label{eq:app-jepa-branches}
\end{equation}
Both branches receive the same observation and interval queries. The online
encoder sees typed, timed absence tokens, whereas the target encoder sees the
complete multimodal sequence. The target states therefore describe how the
same content and task-specific readouts would have been represented under full
evidence.
The target encoder remains in evaluation mode, receives no gradient, and follows
the online parameters through an exponential moving average:
\begin{equation}
\begin{aligned}
\bar{\vtheta}_{t}
&=
\emam_t\bar{\vtheta}_{t-1}
+
(1-\emam_t)\vtheta_t,\\
\emam_t
&=
0.996
+
(0.9999-0.996)
\min\!\left(\frac{t}{T_{\mathrm{tot}}},1\right).
\end{aligned}
\label{eq:app-ema}
\end{equation}
The momentum is scheduled over optimization steps and reaches $0.9999$ at the
end of training. Model buffers are copied from the online encoder rather than
averaged. This EMA and stop-gradient construction provides slowly evolving
contextual targets without requiring negative pairs
~\citep{grill2020byol,assran2023ijepa,baevski2022data2vec}.

A single token-wise predictor is shared by content and query states:
\begin{equation}
\pred(\mathbf{z})
=
W_2
\operatorname{GELU}\!\left(
W_1\operatorname{LN}(\mathbf{z})+\mathbf{b}_1
\right)
+\mathbf{b}_2,
\qquad
\dmod\rightarrow384\rightarrow\dmod.
\label{eq:app-predictor}
\end{equation}
The predictor is a bottleneck rather than an expansion and contains no
attention. It cannot gather evidence across tokens after encoding. Cross-modal
and temporal recovery must therefore occur within the online encoder, which is
the component retained at inference.
The predictor also receives no separate mask-location sequence. Its only
knowledge of missingness comes through the typed, timed absence states already
processed by the encoder. A mask-conditioned or attention-based predictor could
learn to invert the corruption itself, allowing the retained encoder to preserve
outage-specific representations~\citep{garrido2024iwm}. Limiting the predictor
instead encourages stability in the task-conditioned readouts. The content
states may still retain information about which sensor failed.
Teacher targets are detached and normalized using a parameter-free LayerNorm.
For compactness, define
\[
d_1(\mathbf{u},\mathbf{v})
=
\frac{1}{\dmod}\lVert\mathbf{u}-\mathbf{v}\rVert_1,
\qquad
\widehat{\mathbf{z}}
=
\operatorname{LN}\!\left(
\operatorname{sg}[\bar{\mathbf{z}}]
\right).
\]
Layer normalization removes target scale as a trivial degree of freedom, while
the mean absolute distance follows latent-prediction objectives such as
V-JEPA~\citep{assran2025vjepa2}.
Let $v_q\in\{0,1\}$ identify valid, unpadded query entries. The content and query
objectives are
\begin{equation}
\begin{aligned}
\lctx
&=
\frac{1}{\Ncon}
\sum_{i=1}^{\Ncon}
d_1\!\left(
\pred(\vz^{c}_{\vtheta,i}),
\widehat{\vz}^{c}_{i}
\right),\\
\lqry
&=
\frac{
\displaystyle
\sum_{q=1}^{3N_v+N_a}
v_q\,
d_1\!\left(
\pred(\vz^{q}_{\vtheta,q}),
\widehat{\vz}^{q}_{q}
\right)
}{
\displaystyle
\sum_{q=1}^{3N_v+N_a}v_q
},\\
\Ljepa
&=
\lctx+\lqry.
\end{aligned}
\label{eq:app-content-query-jepa}
\end{equation}

The content loss is dense over all $\Ncon=2\Nrow$ states, rather than restricted to
substituted positions. Transformer representations depend on their context, so
a local outage can also alter observed tokens elsewhere in the sequence. Dense
content JEPA aligns this entire corrupted context with its clean counterpart
instead of treating the objective as masked-position reconstruction.
The query term aligns the same task and interval under corrupted and clean
evidence. Content and query losses are normalized separately before summation,
giving the two sets equal aggregate weight despite their different sizes. Since
an observation typically contains fewer query states than content states, an
individual query consequently receives greater average weight. This emphasis is
deliberate because the classifiers operate directly on query representations.
Content alignment cannot replace query alignment. The attention mask prevents
content tokens from reading queries, and therefore
\begin{equation}
\frac{\partial\lctx}{\partial\clsv{h}}=0,
\qquad
\forall h\in\Hset.
\label{eq:app-query-gradient-path}
\end{equation}
The content loss still updates the shared encoder, but it has no direct gradient
path to the learned query seeds. Query JEPA closes this path by constraining the
exact interval-conditioned representation consumed by each classifier. It is
therefore the term that directly refines query attention toward evidence that
survives an outage.

\subsection{Task-Specific Prediction Heads}
\label{app:prediction-heads}

Each classifier reads only its corresponding query representation. Verb, noun,
and sound use independent linear heads
\begin{equation}
C_h(\mathbf{z})
=
W_h\operatorname{Drop}_{0.3}(\mathbf{z})+\mathbf{b}_h,
\qquad
h\in\{\mathrm{verb},\mathrm{noun},\mathrm{sound}\},
\label{eq:app-linear-heads}
\end{equation}
with 97, 300, and 44 outputs, respectively.
For the 3,806 \EK action classes, we use a compositional classifier. If $v(c)$
and $n(c)$ denote the verb and noun associated with action $c$, its score is
\begin{equation}
S_c(\mathbf{z})
=
a_{v(c)}(\mathbf{z})
+
b_{n(c)}(\mathbf{z})
+
I_c,
\qquad
c\in\{1,\ldots,3806\}.
\label{eq:app-compositional-action}
\end{equation}
Here,
$a:\mathbb{R}^{\dmod}\rightarrow\mathbb{R}^{97}$ and
$b:\mathbb{R}^{\dmod}\rightarrow\mathbb{R}^{300}$ belong exclusively to the
action head. They are not shared with the standalone verb and noun classifiers.
The learned residual $I_c$, initialized to zero, captures compatibility specific
to each verb--noun pair.
This factorization shares evidence across the long-tailed action vocabulary.
Rare actions benefit from observations of their constituent verbs and nouns,
while $I_c$ preserves pair-specific information. The four heads otherwise share
no parameters and interact only through the common encoder.

\subsection{VICReg and Complete Objective}
\label{app:complete-objective}

For each head, classification is the mean label-smoothed cross-entropy over its
valid queries, with smoothing $\varepsilon=0.1$. A head with no valid query in a
batch contributes an exact zero. The four task losses are combined as
\begin{equation}
\mathcal{L}_{\mathrm{CE}}
=
\mathcal{L}_{\mathrm{CE}}^{\mathrm{verb}}
+
\mathcal{L}_{\mathrm{CE}}^{\mathrm{noun}}
+
w_{\mathrm{act}}\mathcal{L}_{\mathrm{CE}}^{\mathrm{action}}
+
\lambda_A\mathcal{L}_{\mathrm{CE}}^{\mathrm{sound}},
\qquad
\lambda_A=0.1.
\label{eq:app-classification-loss}
\end{equation}
The sound weight is explicit because visual and acoustic queries have different
counts and label spaces, while the action coefficient controls the contribution
of the larger compositional action vocabulary.

We additionally apply VICReg to observed online content states
\citep{bardes2022vicreg}. Let
$\mathbf{Z}\in\mathbb{R}^{N\times\dmod}$ contain all non-substituted content
states in the batch and let
$\widetilde{\mathbf{Z}}=\mathbf{Z}-\mathbb{E}[\mathbf{Z}]$. The regularizers are
\begin{equation}
\begin{aligned}
\sigma_j
&=
\sqrt{
\operatorname{Var}(\widetilde{\mathbf{Z}}_{:,j})
+10^{-4}
},\\
\mathcal{L}_{\mathrm{var}}
&=
\frac{1}{\dmod}
\sum_{j=1}^{\dmod}
\max(0,1-\sigma_j),\\
\mathbf{C}
&=
\frac{
\widetilde{\mathbf{Z}}^\top
\widetilde{\mathbf{Z}}
}{N-1},
\qquad
\mathcal{L}_{\mathrm{cov}}
=
\frac{1}{\dmod}
\sum_{i\neq j}C_{ij}^{2}.
\end{aligned}
\label{eq:app-vicreg}
\end{equation}

For compactness, we denote the complete weighted VICReg contribution by
\begin{equation}
w_{\mathrm{vic}}\Lvic
\equiv
\mu\mathcal{L}_{\mathrm{var}}
+
\nu\mathcal{L}_{\mathrm{cov}} .
\label{eq:app-weighted-vicreg}
\end{equation}

The variance term prevents individual dimensions from collapsing, while the
covariance term discourages redundant dimensions. Substituted tokens are
excluded because their learned absence embeddings are constant content values.
Including them would force these special tokens to reproduce the statistics of
genuine sensor observations. Both terms are computed in fp32 for numerical
stability.
The complete scheduled objective is

\begin{equation}
\boxed{
\mathcal{L}
=
w_{\mathrm{ce}}\mathcal{L}_{\mathrm{CE}}
+
w_{\mathrm{jepa}}\Ljepa
+
w_{\mathrm{vic}}\Lvic
}
\label{eq:app-total-objective}
\end{equation}
with $\lambda_A=0.1$, and $\nu=1$. The schedules for
$w_{\mathrm{ce}}$, $w_{\mathrm{act}}$, $w_{\mathrm{jepa}}$, $\mu$, and the
clean-observation probability are specified in \Cref{app:training-schedule} and \Cref{tab:app-objective-schedule}. All
objectives are active from the beginning of training and optimized jointly.
MacJEPA never reconstructs raw audio or video. Window-local dropout defines the
unavailable evidence, dense content JEPA stabilizes the resulting multimodal
context, and query JEPA preserves the representation used for prediction. The
EMA teacher and predictor are discarded after training, leaving only the online
encoder and task heads at inference.

\section{Additional Ablations}
\label{app:exp_ex}

Beyond the controlled component analysis in
\Cref{tab:component-analysis}, we examine MacJEPA's sensitivity to its training
and architectural choices. Each variant is trained from scratch for 50 epochs
and evaluated using its final epoch-50 online checkpoint, with only the factor
under study changed. Following the nested per-window temporal missingness protocol defined in
\Cref{sec:experimental_setup,app:evaluation-protocol}, the first four ablations report
annotation-level top-1 accuracy at rates
$r\in\{0,0.25,0.50,0.75,1\}$ for \EK action under video removal and \ES sound
under audio removal, alongside the corresponding unimodal references. We study
the training-time modality-drop probability, outage-window duration,
observation length, and encoder depth, followed by an \EK action ablation of
the frozen visual features. Together, these studies isolate how outage
exposure, temporal geometry, model capacity, and input representation shape
missing-modality robustness

\subsection{Effect of Training-Time Modality-Drop Probability}
\label{app:ablation-drop-rate}

We isolate the window-level corruption probability
\[
\pdrop
=
\Pr\!\left(
\sum_{m\in\Mset}\Mel{k}{m}=1
\right),
\]
where $\Mel{k}{m}=1$ denotes that modality $m$ is removed from window $k$.
Within observations selected for corruption, a window remains complete with
probability $1-\pdrop$ and loses either modality with probability
$\pdrop/2$. We vary only
$\pdrop\in\{0.3,0.5,0.7,0.9\}$, keeping the observation length
$\Nrow=36$, window size $\wsz=4$, six-layer encoder, frozen feature backbones,
clean-observation schedule, learning objectives, and optimization fixed.
MacJEPA uses $\pdrop=0.7$.
\Cref{tab:app-drop-rate} exposes a clear
robustness--availability trade-off. On \EK, increasing $\pdrop$ from $0.3$ to
$0.9$ raises accuracy under complete video loss from 13.7 to 21.5, but reduces
clean and $75\%$-missing accuracy from 54.1 to 53.6 and from 49.7 to 46.5.
Frequent training outages therefore improve reliance on the surviving audio
stream, while excessive corruption weakens the visual evidence available for
learning action composition. The effect is milder on \ES: complete-loss
accuracy rises from 47.8 at $\pdrop=0.3$ to 50.1 at $\pdrop=0.5$ and then
saturates, while clean accuracy remains within 57.4--58.1. We retain $\pdrop=0.7$ since it exposes most eligible windows to an outage while preserving enough complete cross-modal evidence to learn fusion, providing a balanced policy across tasks and missing rates rather than optimizing a single evaluation endpoint.
\begin{table}[t]
\caption{\small Effect of the training-time modality-drop probability
$\pdrop$. We report top-1 accuracy (\%) across per-window missing rates for
\EK action under video removal and \ES sound under audio removal.
$\pdrop=0.7$ is the MacJEPA setting. Follows nested per-window temporal missingness protocol.}
\label{tab:app-drop-rate}
\centering
\footnotesize
\setlength{\tabcolsep}{2pt}
\setlength{\arrayrulewidth}{0.7pt}
\renewcommand{\arraystretch}{1.15}

\begin{tabularx}{\linewidth}{
  @{}>{\raggedright\arraybackslash}X
  *{5}{>{\centering\arraybackslash}p{0.042\linewidth}} |
  *{5}{>{\centering\arraybackslash}p{0.042\linewidth}}@{}
}
\hline
Training setting &
\multicolumn{5}{c|}{\EK action, V drop} &
\multicolumn{5}{c}{\ES sound, A drop} \\
\cline{2-6}
\cline{7-11}
& 0 & 25 & 50 & 75 & 100
& 0 & 25 & 50 & 75 & 100 \\
\hline
\emph{Unimodal reference}
& 13.0 & 13.0 & 13.0 & 13.0 & 13.0
& 41.4 & 41.4 & 41.4 & 41.4 & 41.4 \\
\hline
$\pdrop=0.3$
& \textbf{54.1} & \textbf{53.9} & \textbf{53.6}
& \textbf{49.7} & 13.7
& 57.4 & 56.7 & 55.4 & 53.1 & 47.8 \\
$\pdrop=0.5$
& 53.9 & \textbf{53.9} & 53.2 & 48.8 & 16.9
& 57.7 & \textbf{57.2} & \textbf{55.9}
& \textbf{54.1} & \textbf{50.1} \\
$\pdrop=0.9$
& 53.6 & 53.2 & 51.8 & 46.5 & \textbf{21.5}
& \textbf{58.1} & \textbf{57.2} & 55.8 & 53.6
& \textbf{50.1} \\
\hline
$\pdrop=0.7$ (MacJEPA)
& 54.0 & 53.7 & 52.2 & 47.3 & 19.5
& \textbf{58.1} & 56.0 & 54.3 & 52.1 & 50.0 \\
\hline
\end{tabularx}
\end{table}

\subsection{Effect of Outage Granularity or Window Length}
\label{app:ablation-window-size}

We isolate the number of feature rows $\wsz$ grouped into one outage window.
For a fixed observation of $\Nrow=36$ rows, this determines the number and
duration of independently masked intervals as
\[
\Nwin=\frac{\Nrow}{\wsz},
\qquad
\windursym=\wsz\fstride.
\]
We vary only $\wsz\in\{2,3,4,6\}$, producing outages of
$1.07$, $1.60$, $2.13$, and $3.20$\,s. The observation set,
$\pdrop=0.7$, clean-observation schedule, frozen features, six-layer encoder,
objectives, and optimization remain fixed. MacJEPA uses $\wsz=4$, giving nine
$2.13$\,s windows per observation. At fixed $\Nrow$, changing $\wsz$ also changes
$\Nwin=\Nrow/\wsz$ and hence the $3^{\Nwin}$ possible train-time availability
patterns. Fine windows produce more independently located short outages, while
coarse windows expose the model to fewer but longer failures.
\Cref{tab:app-window-size} shows that outage duration controls the type of
robustness learned while leaving clean recognition comparatively stable. Short
$1.07$\,s windows are easiest to bridge using neighboring temporal and
cross-modal evidence, giving the strongest intermediate robustness. At $75\%$
video missingness, they reach 50.9 on \EK, compared with 44.5 for $3.20$\,s
windows. Longer training outages instead prepare the model for sustained sensor
failure, raising complete-video-loss accuracy from 14.8 to 22.3 as $\wsz$
increases from 2 to 6. The same distinction is weaker but remains visible on
\ES, where short windows lead at intermediate rates and the coarser settings
perform best under complete audio loss. We retain $\wsz=4$ because its
$2.13$\,s outage exceeds the $1.07$\,s feature receptive field while preserving
nine independently addressable intervals. This setting achieves the strongest
clean accuracy on both tasks and balances partial- and complete-outage
robustness without optimizing for either dataset or missingness endpoint.

\begin{table}[t]
\caption{\small Effect of outage granularity $\wsz$. We report top-1 accuracy
(\%) across per-window missing rates for \EK action under video removal and
\ES sound under audio removal. $\wsz=4$ is the MacJEPA setting. Follows nested per-window temporal missingness protocol.}
\label{tab:app-window-size}
\centering
\footnotesize
\setlength{\tabcolsep}{2pt}
\setlength{\arrayrulewidth}{0.7pt}
\renewcommand{\arraystretch}{1.15}

\begin{tabularx}{\linewidth}{
  @{}>{\raggedright\arraybackslash}X
  *{5}{>{\centering\arraybackslash}p{0.042\linewidth}} |
  *{5}{>{\centering\arraybackslash}p{0.042\linewidth}}@{}
}
\hline
Window setting &
\multicolumn{5}{c|}{\EK action, V drop} &
\multicolumn{5}{c}{\ES sound, A drop} \\
\cline{2-6}
\cline{7-11}
& 0 & 25 & 50 & 75 & 100
& 0 & 25 & 50 & 75 & 100 \\
\hline
\emph{Unimodal reference}
& 13.0 & 13.0 & 13.0 & 13.0 & 13.0
& 41.4 & 41.4 & 41.4 & 41.4 & 41.4 \\
\hline
$\wsz=2$ ($18\times1.07$\,s)
& 53.6 & 53.6 & \textbf{53.1} & \textbf{50.9} & 14.8
& 57.7 & 56.9 & \textbf{56.1} & \textbf{54.1} & 48.6 \\
$\wsz=3$ ($12\times1.60$\,s)
& 53.5 & 53.5 & 52.4 & 48.6 & 16.9
& 57.1 & 56.6 & 55.7 & \textbf{54.1} & 49.3 \\
$\wsz=6$ ($6\times3.20$\,s)
& 53.7 & 53.3 & 51.4 & 44.5 & \textbf{22.3}
& 57.9 & \textbf{57.3} & 55.9 & 53.7 & \textbf{50.0} \\
\hline
$\wsz=4$ ($9\times2.13$\,s; MacJEPA)
& \textbf{54.0} & \textbf{53.7} & 52.2 & 47.3 & 19.5
& \textbf{58.1} & 56.0 & 54.3 & 52.1 & \textbf{50.0} \\
\hline
\end{tabularx}
\end{table}

\subsection{Effect of Observation Length}
\label{app:ablation-observation-length}

We isolate the number of feature rows $\Nrow$ in each observation. With feature
stride $\fstride$ and fixed window size $\wsz=4$, the observation span and
number of outage windows are
\[
\obsspan=\Nrow\fstride,
\qquad
\Nwin=\frac{\Nrow}{\wsz}.
\]
We vary only $\Nrow\in\{20,28,36,44\}$, corresponding to observation spans of
$10.67$, $14.93$, $19.20$, and $23.47$\,s. The window size
$\wsz=4$, $\pdrop=0.7$, clean-observation schedule, frozen features,
six-layer encoder, objectives, and optimization remain fixed. Observation
indices are regenerated for each length, while multi-view averaging preserves
annotation-level evaluation with essentially unchanged denominators. MacJEPA
uses $\Nrow=36$.

\begin{table}[t]
\caption{\small Effect of observation length $\Nrow$. We report top-1 accuracy
(\%) across per-window missing rates for \EK action under video removal and
\ES sound under audio removal. $\Nrow=36$ is the MacJEPA setting. Follows nested per-window temporal missingness protocol.}
\label{tab:app-observation-length}
\centering
\footnotesize
\setlength{\tabcolsep}{2pt}
\setlength{\arrayrulewidth}{0.7pt}
\renewcommand{\arraystretch}{1.15}

\begin{tabularx}{\linewidth}{
  @{}>{\raggedright\arraybackslash}X
  *{5}{>{\centering\arraybackslash}p{0.042\linewidth}} |
  *{5}{>{\centering\arraybackslash}p{0.042\linewidth}}@{}
}
\hline
Observation setting &
\multicolumn{5}{c|}{\EK action, V drop} &
\multicolumn{5}{c}{\ES sound, A drop} \\
\cline{2-6}
\cline{7-11}
& 0 & 25 & 50 & 75 & 100
& 0 & 25 & 50 & 75 & 100 \\
\hline
\emph{Unimodal reference}
& 13.0 & 13.0 & 13.0 & 13.0 & 13.0
& 41.4 & 41.4 & 41.4 & 41.4 & 41.4 \\
\hline
$\Nrow=20$ ($10.67$\,s, $\Nwin=5$)
& 53.7 & 53.6 & 52.7 & 47.5 & \textbf{22.0}
& 57.7 & 57.0 & 55.9 & 53.6 & 49.5 \\
$\Nrow=28$ ($14.93$\,s, $\Nwin=7$)
& 53.7 & 53.6 & \textbf{52.9} & \textbf{47.7} & 19.6
& \textbf{58.1} & \textbf{57.6} & \textbf{56.1}
& \textbf{53.9} & 49.9 \\
$\Nrow=44$ ($23.47$\,s, $\Nwin=11$)
& 53.6 & 53.3 & 52.5 & \textbf{47.7} & 18.4
& 56.6 & 56.3 & 55.0 & 53.0 & 49.2 \\
\hline
$\Nrow=36$ ($19.20$\,s, $\Nwin=9$; MacJEPA)
& \textbf{54.0} & \textbf{53.7} & 52.2 & 47.3 & 19.5
& \textbf{58.1} & 56.0 & 54.3 & 52.1 & \textbf{50.0} \\
\hline
\end{tabularx}
\end{table}

Because $\wsz$ remains fixed, varying $\Nrow$ also changes
$\Nwin=\Nrow/\wsz$. At missing rate $r$, the expected outage therefore covers
$r\Nwin$ windows, or $r\obsspan$ seconds, so longer observations provide both
more context and more possible outage locations.
\Cref{tab:app-observation-length} shows that wider temporal context does not
uniformly improve robustness. The $14.93$\,s observation performs best through
most intermediate missing rates, reaching 52.9 and 47.7 on \EK at $50\%$ and
$75\%$, and 57.6, 56.1, and 53.9 on \ES from $25\%$ to $75\%$. This span
provides useful neighboring evidence without introducing as much unrelated
background or additional events. The shortest observation instead leads under
complete video loss on \EK, while extending the context to $23.47$\,s provides
no consistent benefit. Longer context therefore increases available evidence
but also makes attention resolve more temporally distant and potentially
irrelevant content.

We retain $\Nrow=36$ because its $19.2$\,s span captures multiple overlapping
events and nine independently masked intervals while achieving the strongest
clean \EK accuracy, matching the strongest clean \ES result, and leading under
complete audio loss. Shorter observations win isolated missingness columns, but
none provides the same balance between complete-input recognition and
robustness across both tasks.

\subsection{Effect of Encoder Capacity}
\label{app:ablation-encoder-depth}

We isolate the transformer depth $L_{\mathrm{enc}}$ and vary only
$L_{\mathrm{enc}}\in\{2,4,6,8\}$. The model width remains 1,024 with eight
attention heads and an MLP ratio of two. For every depth, drop-path increases
linearly from $0$ to $0.1$, preserving the same terminal regularization rate.
The observation length $\Nrow=36$, window size $\wsz=4$, $\pdrop=0.7$,
clean-observation schedule, frozen features, objectives, and optimization remain
fixed. MacJEPA uses $L_{\mathrm{enc}}=6$.

\Cref{tab:app-encoder-depth} shows no monotonic relationship between capacity
and missingness robustness. Increasing the model from 21.2M to 71.6M
parameters changes clean \EK and \ES accuracy by only 0.3 and 0.2 points,
respectively. The eight-layer model leads at intermediate \EK missingness and
under complete audio loss on \ES, whereas the two-layer model is strongest at
intermediate \ES rates. The six-layer model instead leads clean recognition on
both tasks and complete video-loss recognition on \EK. Even the two-layer
variant retains approximately $99\%$ of its clean accuracy using only $39\%$
of the parameters, indicating that robustness is produced primarily by the
training objective rather than encoder scale. We retain
$L_{\mathrm{enc}}=6$ because it balances capacity and computation while giving
the strongest clean performance on both tasks and competitive robustness across
the full sweep, rather than optimizing an isolated missingness rate.

\begin{table}[t]
\caption{\small Effect of encoder depth $L_{\mathrm{enc}}$. We report top-1
accuracy (\%) across per-window missing rates for \EK action under video
removal and \ES sound under audio removal. Total parameter counts are shown in
parentheses. Follows nested per-window temporal missingness protocol.}
\label{tab:app-encoder-depth}
\centering
\footnotesize
\setlength{\tabcolsep}{2pt}
\setlength{\arrayrulewidth}{0.7pt}
\renewcommand{\arraystretch}{1.15}

\begin{tabularx}{\linewidth}{
  @{}>{\raggedright\arraybackslash}X
  *{5}{>{\centering\arraybackslash}p{0.042\linewidth}} |
  *{5}{>{\centering\arraybackslash}p{0.042\linewidth}}@{}
}
\hline
Encoder setting &
\multicolumn{5}{c|}{\EK action, V drop} &
\multicolumn{5}{c}{\ES sound, A drop} \\
\cline{2-6}
\cline{7-11}
& 0 & 25 & 50 & 75 & 100
& 0 & 25 & 50 & 75 & 100 \\
\hline
\emph{Unimodal reference}
& 13.0 & 13.0 & 13.0 & 13.0 & 13.0
& 41.4 & 41.4 & 41.4 & 41.4 & 41.4 \\
\hline
$L_{\mathrm{enc}}=2$ (21.2M)
& 53.4 & 53.2 & 52.1 & 46.7 & 18.2
& 57.6 & \textbf{57.0} & \textbf{56.2} & \textbf{54.0} & 48.8 \\
$L_{\mathrm{enc}}=4$ (38.0M)
& 53.6 & 53.3 & 51.9 & 47.5 & 19.3
& 57.2 & 56.6 & 55.4 & 53.6 & 50.0 \\
$L_{\mathrm{enc}}=8$ (71.6M)
& 53.7 & 53.5 & \textbf{52.4} & \textbf{47.9} & 18.3
& 57.8 & \textbf{57.0} & 56.0 & 53.8 & \textbf{50.5} \\
\hline
$L_{\mathrm{enc}}=6$ (54.8M; MacJEPA)
& \textbf{54.0} & \textbf{53.7} & 52.2 & 47.3 & \textbf{19.5}
& \textbf{58.1} & 56.0 & 54.3 & 52.1 & 50.0 \\
\hline
\end{tabularx}
\end{table}

\subsection{Robustness across Visual Feature Sources}
\label{app:ablation-visual-features}

We next isolate the frozen visual representation $\featseq{\vmod}$ by ablating the frozen unimodal visual encoder $\uenc{v}$. Let
$\mathbf{o}_{j}$ and $\mathbf{m}_{j}$ denote the 1,024-dimensional Omnivore
Swin-B~\citep{girdhar2022omnivore} and VideoMAE-L~\citep{tong2022videomae} features at row $j$. We compare
\[
\feat{\vmod}{j}
\in
\left\{
\mathbf{o}_{j},\;
\mathbf{m}_{j},\;
[\mathbf{o}_{j}\Vert\mathbf{m}_{j}]
\right\},
\]
giving visual dimensions $\featdim{\vmod}\in\{1024,1024,2048\}$. Only the
visual stem input width changes. Auditory SlowFast~\citep{kazakos2021asf} features, $\Nrow=36$,
$\wsz=4$, $\pdrop=0.7$, the six-layer encoder, clean-observation schedule,
objectives, and optimization remain fixed. MacJEPA concatenates Omnivore and
VideoMAE features.
\Cref{tab:app-visual-features} shows that MacJEPA's robustness is not tied to
one visual backbone. Both single-backbone variants remain well above the
audio-only reference through most of the sweep, while concatenating their
representations improves action accuracy whenever visual evidence remains. It
raises clean accuracy by 1.3 points over Omnivore and 1.5 points over VideoMAE,
with gains of 2.0 and 3.9 points at $75\%$ video missingness. Under complete
video loss, all variants converge within 1.3 points because their visual
features are no longer available, and Omnivore's 0.1-point advantage over the
combined model is marginal. We therefore retain the concatenated
representation because it provides the best balance of clean and partial-outage
recognition while preserving near-best complete-loss robustness, rather than
selecting a visual backbone from an endpoint at which vision is entirely
absent.

\begin{table}[t]
\caption{\small Robustness across frozen visual feature sources. We report
\EK action top-1 accuracy (\%) under per-window video removal. Omnivore and
VideoMAE are concatenated in the MacJEPA setting.  Follows nested per-window temporal missingness protocol.}
\label{tab:app-visual-features}
\centering
\footnotesize
\setlength{\tabcolsep}{5pt}
\setlength{\arrayrulewidth}{0.7pt}
\renewcommand{\arraystretch}{1.15}

\begin{tabularx}{\linewidth}{
  @{}>{\raggedright\arraybackslash}X
  *{5}{>{\centering\arraybackslash}c}@{}
}
\hline
Visual features &
\multicolumn{5}{c}{Video missing rate (\%)} \\
\cline{2-6}
& 0 & 25 & 50 & 75 & 100 \\
\hline
\emph{Unimodal reference (A only)}
& 13.0 & 13.0 & 13.0 & 13.0 & 13.0 \\
\hline
Omnivore only ($\featdim{\vmod}=1024$)
& 52.7 & 52.4 & 51.2 & 45.3 & \textbf{19.6} \\
VideoMAE only ($\featdim{\vmod}=1024$)
& 52.5 & 51.7 & 49.6 & 43.4 & 18.3 \\
\hline
Omnivore $+$ VideoMAE ($\featdim{\vmod}=2048$; MacJEPA)
& \textbf{54.0} & \textbf{53.7} & \textbf{52.2}
& \textbf{47.3} & 19.5 \\
\hline
\end{tabularx}
\end{table}

\section{Extended Implementation Details}
\label{app:implementation}

We provide the evaluation protocol, complete data deatils, feature extraction, sequence construction,
architecture, and training used for every reported result. For all experiments, we set
$(\fstride,\fextent,\Nrow,\wsz,\dcon,\dtim,\ovl,\pdrop)
=(16/30\,\mathrm{s},32/30\,\mathrm{s},36,4,768,256,0.2\,\mathrm{s},0.7)$,
giving
$(\obsspan,\Nwin,\windursym,\Ncon,\dmod)
=(19.2\,\mathrm{s},9,2.13\,\mathrm{s},72,1024)$.

\subsection{Evaluation Protocol}
\label{app:evaluation-protocol}

We evaluate the epoch-50 online model rather than the EMA teacher. Observations
advance by one feature row, so each annotation appears in multiple overlapping
views. Following \citet{chalk2024tim}, we average its logits across all views
before computing annotation-level top-$k$ accuracy.
We use two missingness protocols with identical clean ($r=0$) and complete-loss
($r=1$) endpoints.

\textbf{Nested per-window temporal missingness.} 
Controlled analyses in
\Cref{subsec:design-hypotheses,app:exp_ex} use the nested per-window temporal missingness protocol
designed to test robustness to local sensor disappearance and recovery. Each
observation--window pair receives a fixed pseudorandom rank, thresholded at
$r\in\{0,0.25,0.50,0.75,1\}$, so windows removed at one rate remain removed at
higher rates. At $r=0.75$, the selected modality is absent from 6.75 of nine
windows, or 14.4\,s, on average. Overlapping observations receive independent
ranks, so a physical moment is removed from approximately a fraction $r$ of
its views.

\textbf{Observation-level missingness.}
For comparison with prior work in \Cref{sec:exp:robust}, we follow the
observation-level missingness protocol of \textsc{MMT} and \textsc{MiDl}: a fraction $r$ of
validation observations loses the selected modality for the full $19.2$\,s,
while the remainder stay complete
~\citep{ramazanova2025mmt,ramazanova2025midl}. For fixed MacJEPA, this
corresponds to the endpoint mixture
$\mathrm{Acc}(r)=(1-r)\mathrm{Acc}(0)+r\mathrm{Acc}(1)$. Thus, intermediate
rates vary outage incidence across observations rather than outage duration
within them.

\textsc{MMT} and \textsc{MiDl} classify one annotation from a trimmed 2\,s single-event clip
and evaluate \EK and \ES with separate models, whereas our
model jointly queries both annotation sets within each 19.2\,s untrimmed
observation containing multiple overlapping audio-visual events. Each forward pass contains 8.59 annotations on average, comprising
4.55 \EK and 4.04 \ES annotations. \EK targets occupy a mean/median
16.7\%/9.8\% of the observation, compared with 21.4\%/6.4\% for \ES, whose
median event spans only about 1.2\,s. Despite fewer queries, \ES presents the
denser attribution problem: 49.4\% temporally overlap another sound event,
compared with 40.5\% on \EK. Since query-to-query attention is blocked, each
query must resolve its target independently from shared content. Longer context aids outage
recovery but introduces distractor suppression and multi-instance attribution
absent from clip-based evaluation. Thus, \textsc{TIM} matches our observation
protocol, while \textsc{MMT} and \textsc{MiDl} match our observation-level missingness protocol.
No prior reference matches both.

Missing video and audio are evaluated separately. Video is dominant for \EK and
audio for \ES, with the reverse defining auxiliary-modality removal. We report
annotation-level top-1 accuracy. All \EK heads share one valid-annotation set,
and action accuracy uses the compositional head's own argmax rather than the
conjunction of separate verb and noun predictions.

\subsection{Dataset}
\label{app:data}

\EK-100 contains 100 hours of unscripted, head-mounted recordings from 45
kitchens, comprising 700 untrimmed videos, approximately 20 million frames, and
90,000 fine-grained action segments
~\citep{damen2022rescaling}. Each action is annotated by its temporal extent,
verb, and noun. The label space contains 97 verbs, 300 nouns, and 3,806
verb--noun action classes present across the training and validation sets.
\ES adds temporally localized acoustic annotations to the same underlying
videos~\citep{huh2023epicsounds}. It contains 78.4K categorized audible events
across 44 classes, together with 39.2K non-categorized sound segments. We use
only the categorized training and validation annotations. The acoustic
intervals and labels are defined independently of the visual actions, so the two
annotation streams may differ in duration, overlap, or occur without a
corresponding event in the other modality.
We use the \EK and \ES annotations distributed with \TIM, preserving its
video-disjoint splits and class numbering~\citep{chalk2024tim}. We evaluate 9,668 \EK and 8,045
\ES validation annotations.

\subsection{Indexing and Observation Construction}
\label{app:indexing}

Each observation contains 36 feature rows and advances by one row, or
$16/30$\,s, along the untrimmed video. Consecutive observations consequently
overlap by 35 rows. An annotation is assigned when its overlap with the
observation is at least $0.2$\,s or when it is fully contained within the
observation, following ~\citet{chalk2024tim}. Its interval is clipped to the
observation and normalized by the 19.2\,s observation duration.
Training observations contain, on average, 5.48 \EK and 5.28 \ES queries;
validation observations contain 4.55 and 4.04, respectively. Visual and acoustic queries are padded independently, with
validity masks excluding padded entries from classification, query JEPA, and
multi-view accumulation. Our feature-complete
index contains 487 training and 136 validation videos, producing 485,786
training and 83,964 validation observations.

\subsection{Frozen Feature Extraction}
\label{app:feature-extraction}

All feature backbones are frozen and run offline. Omnivore Swin-B is pretrained on ImageNet,
Kinetics, and SUN RGB-D
~\citep{girdhar2022omnivore,russakovsky2015imagenet,kay2017kinetics,
song2015sunrgbd}. The VideoMAE-L branch follows VideoMAE/InternVideo pretraining
on Kinetics, Something-Something V2, AVA, and WebVid2M
~\citep{tong2022videomae,wang2022internvideo,kay2017kinetics,
goyal2017something,gu2018ava,bain2021frozen}. Both visual extractors are then
fine-tuned on \EK-100 labels. Auditory SlowFast R50 is initialized from
VGGSound and fine-tuned on \ES labels
~\citep{kazakos2021asf,chen2020vggsound,huh2023epicsounds}. Further pretraining,
fine-tuning, and extraction details are provided in the respective extractor
papers ~\citep{girdhar2022omnivore,tong2022videomae,wang2022internvideo,
kazakos2021asf}.
Each visual row concatenates 1,024-dimensional Omnivore and VideoMAE features,
forming a 2,048-dimensional input. Omnivore samples 32 frames, resizes the short
side to 256 pixels, and takes a $224\times224$ center crop. VideoMAE samples
16 frames using a $224\times224$ crop. Both apply ImageNet normalization, and
their outputs are concatenated without learned fusion.
Audio rows use 2,304-dimensional Auditory SlowFast features. Audio is sampled
at 24\,kHz, with 24,000 samples per feature and a 12,800-sample hop that
matches the $\fstride=16/30$\,s visual-grid stride. The log-mel input contains
128 frequency bins and 200 temporal frames, using a 10-ms analysis window and
5-ms hop. Auditory SlowFast uses a slow--fast ratio of four.
All streams are extracted on the same temporal grid. Each row is assigned the
encoded support $\fextent=32/30$\,s, producing $50\%$ overlap between adjacent
rows. Although the physical audio support is 1.0\,s, the common encoded extent
gives corresponding audio and visual rows identical temporal coordinates. This
compact grid represents 19.2\,s of context with 72 content tokens.
\Cref{tab:app-feature-extraction} summarizes the extraction settings.

\begin{table}[t]
\caption{\small Frozen feature-extraction settings.}
\label{tab:app-feature-extraction}
\centering
\scriptsize
\setlength{\tabcolsep}{4pt}
\renewcommand{\arraystretch}{1.12}
\begin{tabularx}{\linewidth}{@{}lXXX@{}}
\toprule
& Omnivore Swin-B & VideoMAE-L & Auditory SlowFast R50 \\
\midrule
Output width
& 1,024 & 1,024 & 2,304 \\
Input support
& 32 frames & 16 frames, tubelet 2 & 1.0\,s waveform \\
Input size
& $224\times224$ & $224\times224$ & $128\times200$ log-mel \\
Preprocessing
& ImageNet normalization
& ImageNet normalization
& 24-kHz mono audio \\
Extraction precision
& fp16 autocast & fp16 autocast & fp32 \\
\bottomrule
\end{tabularx}
\end{table}

\subsection{Temporal Encoding and Sequence Assembly}
\label{app:sequence-implementation}

Each 19.2\,s observation contains 36 rows grouped into nine four-row windows.
Every 2.13\,s window contributes four visual and four audio tokens, giving 72
content positions. Content row $j$ uses the normalized interval
$[j\fstride,j\fstride+\fextent]/\obsspan$. Its endpoint is not clipped and the
final row reaches $37/36$. Query intervals are clipped to the observation and
normalized to $[0,1]^2$.
A shared TimeMLP with widths
$2\rightarrow256\rightarrow256\rightarrow256$ uses three ReLU layers followed
by LayerNorm. It encodes all content and query intervals and is the model's only
positional signal. Visual and audio stems independently apply feature dropout
of $0.5$, project to 768 dimensions, and apply GELU and LayerNorm. Concatenating
the projected feature with its 256-dimensional time encoding produces a
1,024-dimensional token, after which a learned modality embedding is added.
Mask substitution occurs after the stem and before temporal concatenation. Each
modality has a learned 768-dimensional absence vector initialized from a
truncated normal distribution with standard deviation $0.02$. Within each
window, four visual tokens precede four audio tokens. The flattened sequence is
ordered as
\[
[\text{72 content}\mid
 \text{$N_v$ verb}\mid
 \text{$N_v$ noun}\mid
 \text{$N_v$ action}\mid
 \text{$N_a$ sound}].
\]
Four learned 768-dimensional query seeds, initialized with the same scale,
represent verb, noun, action, and sound. Visual-task queries reuse the visual
modality embedding, while sound queries reuse the audio embedding.

\subsection{Encoder, Attention, and Task Readouts}
\label{app:architecture-implementation}

The encoder contains six pre-normalized transformer blocks of width 1,024, with
eight attention heads and a 2,048-dimensional GELU MLP. Attention and MLP
dropout are $0.3$. Drop-path increases linearly from $0$ to $0.1$ across the
blocks and is applied to both residual branches. A final LayerNorm produces the
encoded states, and sequence-level dropout is set to zero.
Every sequence row can attend to all 72 content columns. Query columns are
visible only to themselves. Content therefore attends globally to content but
not to queries, while each query attends to all content and itself but not to
other queries. Windows are flattened before encoding and impose no attention
boundaries.
Verb, noun, and sound readouts use dropout of $0.3$ followed by independent
linear classifiers with 97, 300, and 44 outputs. The compositional action head
in \Cref{app:prediction-heads} predicts 3,806 action classes. The four heads
share no parameters. The encoder contains 50.40M parameters, the modality stems
3.35M, the heads 0.87M, and the TimeMLP 0.13M. The complete inference model has
54.76M parameters.
Content and query JEPA share one training-only predictor consisting of LayerNorm
and a $1024\rightarrow384\rightarrow1024$ GELU MLP without attention. It adds
0.79M parameters and is optimized jointly with the online model. The EMA
teacher is a non-trainable copy kept in evaluation mode. The predictor and
teacher are discarded at inference.

\subsection{Mask Sampling and Objective Schedule}
\label{app:training-schedule}

Training masks are generated independently for each observation and epoch using
the global seed, epoch, video row, and observation start. The probability of
retaining an observation entirely clean increases from $0.15$ to $0.20$ during
the first five epochs. In every remaining observation, each window independently
drops one uniformly selected modality with probability $0.7$. Audio and video
are never removed simultaneously.
Classification uses label
smoothing $\varepsilon=0.1$ and sound weight $\lambda_A=0.1$. Content and query
JEPA use $\lambda_q=1$. VICReg is applied only to observed content states and
computed in fp32 with variance target $\gamma=1$, numerical constant
$10^{-4}$, and covariance weight $\nu=1$.
All objectives are optimized jointly from the first step. Their coefficients
reach their final values during the first five epochs, as listed in \Cref{tab:app-objective-schedule}. The action-specific ramp limits the early contribution of the 3,806-way action
head. There is no separate self-supervised stage or frozen-encoder probe.

\begin{table}[t]
\caption{\small Training-objective schedule.}
\label{tab:app-objective-schedule}
\centering
\scriptsize
\setlength{\tabcolsep}{6pt}
\renewcommand{\arraystretch}{1.10}
\begin{tabular}{@{}c c c c c c@{}}
\toprule
Epoch & $w_{\mathrm{ce}}$ & $w_{\mathrm{act}}$
& $w_{\mathrm{jepa}}$ & $\mu$ & $p_{\mathrm{clean}}$ \\
\midrule
1      & 0.20 & 0.33 & 2.20 & 4.40 & 0.150 \\
2      & 0.40 & 0.67 & 1.90 & 3.80 & 0.163 \\
3      & 0.60 & 1.00 & 1.60 & 3.20 & 0.175 \\
4      & 0.80 & 1.00 & 1.30 & 2.60 & 0.188 \\
5--50  & 1.00 & 1.00 & 1.00 & 2.00 & 0.200 \\
\bottomrule
\end{tabular}
\end{table}

\subsection{Optimization and Regularization}
\label{app:optimization}

We train every model for 50 epochs with AdamW using
$\beta_1=0.9$, $\beta_2=0.95$, weight decay $10^{-4}$, batch size 256, and
gradient clipping at 1.0 over the online model and predictor. The training index
contains 1,897 full batches per epoch with \texttt{drop\_last=True}, giving
94,850 optimization steps.
The learning rate warms linearly from $10^{-6}$ to $3\times10^{-4}$ over the
first 3,794 steps, corresponding to two epochs. It then follows cosine decay to
$10^{-6}$ without a constant-rate hold period. Training uses fp16 autocast and
gradient scaling, while VICReg is explicitly evaluated in fp32. Feature dropout
is $0.5$, encoder and head dropout are $0.3$, and stochastic depth increases
from $0$ to $0.1$.
The EMA momentum increases linearly from $0.996$ to $0.9999$ over all 94,850
steps. Teacher buffers are copied from the online branch rather than averaged.
We use seed 0 and eight persistent data-loading workers with pinned memory and
shuffled training observations. Training completed without non-finite batches.
All reported full-modality, robustness, component, and additional-ablation
results use the final epoch-50 online checkpoint. The same checkpoint evaluates
\EK and \ES, and no early stopping or task-specific checkpoint selection is
used.